\documentclass{article}

\PassOptionsToPackage{numbers,compress,sort}{natbib}
\usepackage{natbib}

\usepackage{wrapfig}
\usepackage[utf8]{inputenc} %
\usepackage[T1]{fontenc}    %
\usepackage{url}            %
\usepackage{booktabs}       %
\usepackage{longtable}
\usepackage{amsfonts}       %
\usepackage{nicefrac}       %
\usepackage{microtype}      %
\usepackage{xcolor}         %
\usepackage{amsmath} 
\usepackage{graphicx}
\usepackage[rightcaption,raggedright]{sidecap}
\usepackage{subcaption}
\usepackage{multirow}
\usepackage{todonotes}
\usepackage{float}
\usepackage{adjustbox}
\usepackage{placeins}

\usepackage[a4paper,margin=1in]{geometry}
\usepackage{lmodern}        %

\definecolor{priorlight}{RGB}{247,248,250} %
\usepackage[colorlinks=true,linkcolor=black,urlcolor=blue]{hyperref}
\usepackage[most]{tcolorbox}
\tcbset{
  colback=priorlight,
  colframe=black!10,
  boxrule=0pt,
  arc=6pt,
  left=16pt,right=16pt,top=14pt,bottom=14pt,
  enhanced,
  drop shadow={black!15!white}
}

\newcommand{\field}[2]{\textbf{#1:}\enspace #2\par}

\begin{document}

\begin{tcolorbox}
  \hspace*{-0.2cm}\includegraphics[width=0.2\linewidth]{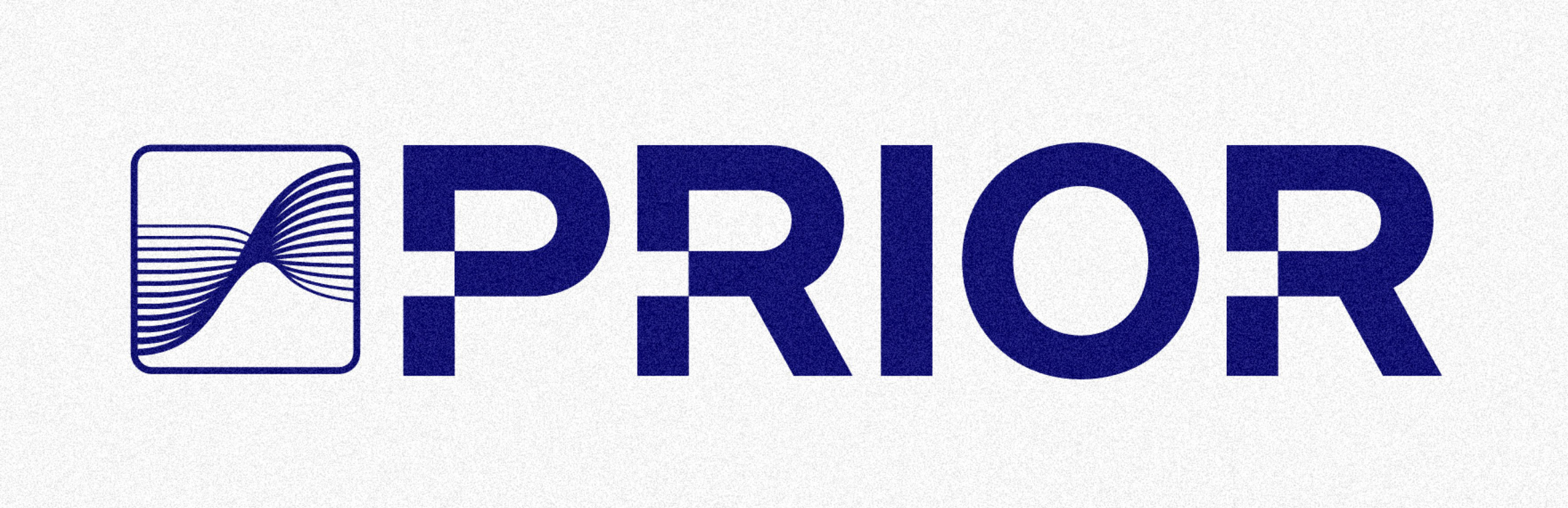}

  {\LARGE\bfseries TabPFN-2.5: Advancing the State of the Art in Tabular Foundation Models\par}

  \vspace{0.35em}
  {\large \hyperref[app:contributors]{Prior Labs Team}$^{1}$\par}

  \vspace{0.25em}
  {\footnotesize $^{1}$The list of contributors can be found in the \hyperref[app:contributors]{appendix}.\par}

  \vspace{0.9em}
  \noindent
  The first tabular foundation model, TabPFN, and its successor TabPFNv2 have impacted tabular AI substantially, with dozens of methods building on it and hundreds of applications across different use cases. \\
This report introduces TabPFN-2.5, the next generation of our tabular foundation model, built for datasets with up to 50{,}000 data points and 2{,}000 features, a $20\times$ increase in data cells compared to TabPFNv2. TabPFN-2.5 is now the leading method for the industry standard benchmark TabArena (which contains datasets with up to 100{,}000 training data points), substantially outperforming tuned tree-based models and matching the accuracy of AutoGluon 1.4, a complex four-hour tuned ensemble that even includes the previous TabPFNv2. 
  Remarkably, default TabPFN-2.5 has a 100\% win rate against default XGBoost on small to medium-sized classification datasets ($\le$10{,}000 data points, 500 features) and a 87\% win rate on larger datasets up to 100K samples and 2K features (85\% for regression).

For production use cases, we introduce a new distillation engine that converts TabPFN-2.5 into a compact MLP or tree ensemble, preserving most of its accuracy while delivering orders-of-magnitude lower latency and plug-and-play deployment.

This new release will immediately strengthen the performance of the many applications and methods already built on the TabPFN ecosystem.

  \vspace{0.8em}
  \field{Date}{November 6, 2025}
  
    \field{Website}{\url{https://priorlabs.ai/}}
    \field{Docs}{\url{https://docs.priorlabs.ai/overview}}
    \field{PyPI}{\texttt{pip install tabpfn-client} (cloud SDK) or \texttt{pip install tabpfn} (local package)}
    \field{License}{\texttt{TABPFN-2.5 License v1.0} (see Section \ref{sec:license} for details)}
    \field{Contact}{hello@priorlabs.ai}
\end{tcolorbox}

\begin{figure}[htbp]
    \centering
    \includegraphics[width=\linewidth]{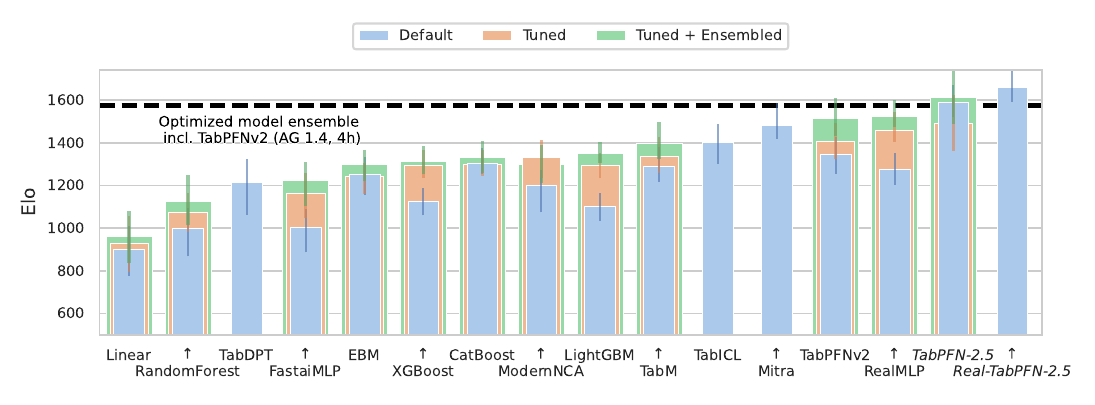}
    \caption{TabPFN-2.5 performance on the standard TabArena-lite benchmark \citep{erickson2025tabarena}, TabPFNv2 classification subset. TabPFN-2.5 outperforms any other model in a forward pass, and marks a strong leap from TabPFNv2. When fine-tuned on real data, Real-TabPFN-2.5 shows even stronger performance. The horizontal dotted line stands for AutoGluon 1.4 extreme mode tuned for 4 hours, an ensemble of models including TabPFNv2. %
    }
    \label{fig:tabarena-hero-plot}
\end{figure}

\section{Introduction}

Tabular data is ubiquitous, forming the backbone of decision-making in countless domains, from finance to healthcare. For decades, traditional tabular machine learning—built on gradient-boosted trees~\citep{chen2016xgboost,prokhorenkova2018catboost,lightgbm}, random forests~\citep{breiman2001random}, and linear or additive models—has been the workhorse of applied data science.  
Yet these methods remain limited: they require extensive dataset-specific tuning, often provide uncalibrated or unreliable uncertainty estimates without significant modification, and lack the generalization and transferability of modern foundation models.

Tabular foundation models (TFMs) offer a new paradigm. %
They address these limitations by pretraining on large synthetic distributions of tabular tasks and performing inference via in-context learning instead of gradient descent. They are training-free predictors meta-trained to yield strong calibration, without the need for time-consuming and labor-intensive hyperparameter tuning necessary for gradient-boosted trees. Their strong generalization makes them particularly attractive for data-scarce domains.

Our initial release, TabPFNv1 \citep{hollmann2022tabpfnv1} served as a proof-of-concept that a transformer could learn a Bayesian-like inference algorithm, though it was limited to small (up to 1k samples), clean, numerical-only data.
Our successor, TabPFNv2 \citep{Hollmann2025tabpfnv2}, scaled this idea into a practical model for datasets up to 10{,}000 samples. TabPFNv2 handles the messy and heterogeneous data seen in the real world—including categorical features, missing values, and outliers.

This paper describes the next release of TabPFN: TabPFN-2.5. 
Our key contributions are: 
\begin{itemize} 
\item \textbf{SOTA Performance:} In a forward pass, TabPFN-2.5 outperforms tuned tree-based models (like XGBoost and CatBoost) and matches the accuracy of AutoGluon 1.4 tuned for 4 hours—a complex ensemble that includes all previous methods, even TabPFNv2. %
\item \textbf{Improved Scalability:} We scale the power of in-context learning to datasets of up to 50{,}000 samples (5x increase over TabPFNv2) and 2{,}000 features (4x increase), making TFMs viable for a much wider range of real-world problems \footnote{In exploratory runs, classification datasets up to $\sim$160k rows $\times$ 500 features and regression datasets up to $\sim$85k $\times$ 500 features fit into memory on an NVIDIA H100~(80\,GB) using FP16 and FlashAttention-3. These configurations are outside our validated range and not included in reported benchmarks.}. While TabPFN-2.5 was designed for up to 50,000 rows, we note that this limit is not strict and report strong results on benchmarks with up to 100,000 training samples.
\item \textbf{Fast Inference:} We dramatically improve inference speed. We introduce TabPFN-as-MLP/TreeEns, a proprietary output engine, that yields an MLP or tree ensemble, combining most of TabPFN's 
accuracy with the low-latency inference and easy deployment of MLPs and tree ensembles. %
\end{itemize}

We begin by surveying the growing ecosystem of TabPFN applications and extensions (Section \ref{sec:usecases_extensions}). We then describe our methodological advances (Section \ref{sec:methods}) and present the experimental results (Section \ref{sec:results}). We then discuss how to get the best speed out of TabPFN on common hardware (Section \ref{sec:speed-common-gpus}) as well as our non-commercial open-source license (Section \ref{sec:license}).
We conclude by discussing the remaining limitations and opportunities for future work (Sections \ref{sec:conclusions}). For installation and usage examples, see the online documentation at
\url{https://docs.priorlabs.ai/}.

\section{Ecosystem \& Adoption} \label{sec:usecases_extensions}

We now discuss community adoption, methods built on top of TabPFN, and our TabPFN extensions.

\subsection{Community Adoption}
Since its release, TabPFNv2 has become a widely used baseline for tabular ML. The \textit{Nature} paper~\cite{Hollmann2025tabpfnv2} has been cited in almost 400 papers within 10 months of its publication, and the open-source package has surpassed 2{,}000{,}000 downloads on PyPI\footnote{Google Scholar entry (accessed Nov 6, 2025; Download stats for \texttt{tabpfn} on pepy.tech (accessed Nov 6, 2025): \url{https://www.pepy.tech/project/tabpfn}}. Adoption spans both research and production, especially in settings with sparse data or frequent retraining requirements. This widespread adoption has matured TabPFN from a research model into a stable product. With feedback from our community of nearly 1,500 users on Discord, and hundreds of closed GitHub issues, we have shipped numerous stability fixes, and cross-platform device compatibility. In addition to commercial use-cases, we also collected 100 published use cases across a broad range of areas (please see Appendix \ref{app:use_cases} for a detailed list):
\begin{itemize}
    \item \textbf{Healthcare and Life Sciences.} Adoption is strongest in healthcare (50+ published applications), driven by TabPFN's exceptional performance in data-scarce settings---a common challenge in medicine. Use cases span oncology, neurology, cardiology, and pharmacology, powering applications like diagnosis, prognosis, and treatment response prediction from complex multimodal (clinical, imaging, omics) data.
    \item \textbf{Financial Services, Banking, and Insurance.} While we see strong commercial traction, public-facing use cases are rare due to the competitive, private nature of this industry (3 collected). Applications in this domain typically involve proprietary forecasting, uplift modeling, and risk assessments.
    \item \textbf{Energy and Utilities.} We've identified 15 published cases centered on complex forecasting and optimization. Key applications include environmental forecasting (algal blooms, wildfire risk), renewable-energy nowcasting, and process/asset optimization across water, oil \& gas.
    \item \textbf{Manufacturing and Industrial.} The 12 diverse published use cases in this area highlight TabPFN's flexibility. Applications include anomaly detection in IIoT security, predictive maintenance for rotating machinery, physics-aware optimization for battery thermal modeling, and semiconductor test optimization.
    \item \textbf{Other Industries} 19 further applications demonstrate broad utility, spanning geoscience, agriculture, materials, and engineering. These range from microbiome classification and lunar regolith analysis to soil property modeling, fuel-blend optimization and crop yield forecasting.
\end{itemize}    

\subsection{A Foundational Layer for New Research} Beyond direct application, TabPFN now serves as a foundational layer for new research domains. Its ability to act as a powerful, pre-trained ``algorithm-in-a-box'' has unlocked new approaches to complex problems. We expect TabPFN-2.5 to directly boost performance in all these areas:

\begin{itemize}
    \item \textbf{Time Series Forecasting}: TabPFN-TS~\cite{hoo2024tabpfn_ts} extends TabPFN to time-series forecasting by incorporating temporal context into its in-context learning mechanism, outperforming specialized time-series models without any retraining.
    
    \item \textbf{Node Classification in Graphs:} Various works~\cite{Hayler2025GraphsTablesZeroShot, eremeev2025turningtabularfoundationmodels} represent graph nodes as tabular instances with relational and structural features, directly using tabular foundation models like TabPFN to solve the problem.

    \item \textbf{Data Streams:} 
    TabPFNv2 was used for in-context learning on Evolving Data Streams~\cite{Lourenco2025ICLStreams}. TabPFN can \emph{adapt to non-stationary data streams} online, without retraining, enabling continual learning in evolving environments.
    
    \item \textbf{Reinforcement Learning:}
    TabPFNv2 was used to replace gradient-based policy optimization with in-context optimization over trajectories, creating a powerful general-purpose optimizer for RL tasks \citep{Schiff2025TabPFNRL}.
    
    \item \textbf{Bayesian optimization: }
    GIT-BO~\cite{Yu2025GITBO} uses TabPFNv2 inside of high-dimensional Bayesian Optimization, as it enables efficient search in high-dimensional and heterogeneous design spaces.
    
    \item \textbf{Multimodal Learning \& Encoding:} TabPFN is used to integrate tabular data with other modalities. It can serve as a \emph{frozen tabular encoder} to generate robust embeddings for combination with data like images (e.g., in the TIME framework~\cite{luo2025timetabpfnintegratedmultimodalengine}), or handle modalities in a unified manner by adding modality-specific projectors~\cite{Lourenco2025ICLStreams}.

    \item \textbf{Causal Inference:} Do-PFN \cite{robertson_dopfn}, CausalPFN \cite{balazadeh_causalpfn},  and CausalFM \cite{feuerriegel_causalfm} pre-train PFNs to predict interventional outcomes, and show strong performance in estimating causal effects.

\end{itemize}

\subsection{The TabPFN-Extensions Ecosystem}
We maintain the TabPFN–Extensions repository (\url{https://github.com/PriorLabs/tabpfn-extensions}), which offers extensions around the core model, developed together with a growing community around TabPFN. These extensions leverage TabPFN capabilities for:
\begin{itemize}
  \item \textbf{Interpretability.} SHAP values, feature selection, partial dependence.
  \item \textbf{Unsupervised Tasks.} Data generation, augmentation, outlier detection.
  \item \textbf{Advanced Modeling.} Many-class classification, regression-via-classifier.
  \item \textbf{Performance \& Integration.} Lightweight HPO, ensembling, and integration with tree/forest baselines.
\end{itemize}
Figure~\ref{fig:workflow} in the appendix provides a minimal workflow to help users pick the right components for their task.

\section{Model Overview}\label{sec:methods}
TabPFN-2.5 follows the same general design as TabPFNv2 but introduces deeper architectures, richer synthetic priors, and new calibration and inference modules. We summarize only the key changes here.

\paragraph{Data.}
We improved our prior data generation substantially, broadened the set of distributions and scaled up to more data points and more features, while keeping the prediction tasks difficult. Like the original TabPFNv2, TabPFN-2.5 is trained purely on synthetically generated data. 
We also release a version that is fine-tuned on real data following Real-TabPFN~\cite{garg2025realtabpfn}. It is trained on a curated corpus of \textbf{~43 real-world tabular datasets} sourced from OpenML and Kaggle, deduplicated against all internal benchmarks and the full TabArena suite. We refer to this version as Real-TabPFN-2.5, and report strong improvement in Figures \ref{fig:tabarena-classification-regression-10k-500} and \ref{fig:tabarena-classification-regression}. See Appendix \ref{data_contamination_real_tabpfn} for details on training and deduplication.

\paragraph{Architecture.}
We follow the alternating-attention transformer design of TabPFNv2, which attends across both data points and features to achieve permutation invariance, but introduces some changes:
\begin{itemize}
    \item We increase the network depth from 12 to 18 layers for our regression model and 24 layers for our classification model.
    \item We simultaneously increase the feature group size (the number of features being embedded together), which allows for faster training and inference. We use a group size of 3 for TabPFN-2.5, compared to 2 for TabPFNv2. 
    \item For our regression models, we found a small improvement by replacing the linear encoder used in TabPFNv2 by a 2-layer MLP.
    \item Finally, we add 64 additional ``thinking'' rows to the input dataset of TabPFN-2.5, which are learned during pretraining. Inspired by results from the LLM literature \citep{MerrillSabharwal2025_ExactExpressivePower, GoyalEtAl2024_ThinkBeforeYouSpeak}, these rows give additional computational capacity to the model and can also act as attention sinks to help the model ignore other rows \citep{DarcetEtAl2024_VisionTransformersNeedRegisters}. 
\end{itemize} Other core components from TabPFNv2—feature/sample dual attention, caching separation of training/test context, and positional feature embeddings—remain unchanged.

\paragraph{Preprocessing.}
We aggregate predictions across multiple dataset permutations and feature transformations to enhance robustness and generalization. 
In the updated TabPFN-2.5 configuration, additional feature transformations are introduced to enhance robustness against outlier-prone feature distributions and to increase the diversity among the individual estimators. Specifically, we combine robust scaling and soft clipping (following \citep{holzmuller2024realmlp}) with quantile transformations and standard scaling to balance stability and sensitivity across features. Following TabPFNv2, we also include singular value decomposition (SVD) components as additional features in some of the estimators, capturing high-energy directions of variance that provide complementary global structure information.

\paragraph{Hyperparameter Tuning of TabPFN with TabPFN.}

TabPFN’s hyperparameter space spans architectural, training, and prior-data parameters, making exhaustive grid search computationally infeasible. To explore this space efficiently, we adopted a surrogate-based optimization strategy.

We first trained $\approx100$ models on a broad but sparse grid of hyperparameter configurations drawn from plausible prior ranges and evaluated them on a curated in-house validation suite, producing a compact set of hyperparameter–performance pairs.

With $\sim50$ hyperparameters and only $100$ datapoints, direct interpolation was prone to overfitting. We therefore used a regression model well-suited for data-scarce structured prediction—our previous TabPFNv2 model—as a surrogate to predict validation performance over a denser grid of $10{,}000$ configurations. This self-referential “TabPFN-tunes-TabPFN” strategy efficiently surfaced promising regions of the search space for full, compute-intensive training runs.

\paragraph{Tuning custom metrics.}
TabPFN-2.5 adds new post-processing capabilities that enhance both calibration and metric-specific optimization. Our framework now supports tuning the classifier’s decision threshold, enabling direct optimization of metrics beyond accuracy—such as the F1-score—by adjusting the operating point to the desired trade-off between precision and recall. For multiclass classification, it allows to apply temperature scaling to the final softmax outputs to improve probability calibration. This threshold tuning procedure can yield substantial performance improvements (see Appendix \ref{app:threshold_tuning}). Unless otherwise noted, however, all classification results in this report are computed using uncalibrated, default scores, without temperature scaling or threshold tuning.\label{app:threshold_tuning}

\paragraph{Reducing inference costs.}
Despite being a larger model than TabPFNv2, TabPFN-2.5 is between 1x and 2.3x faster thanks to optimized preprocessing and larger feature groups, as shown in Figure~\ref{fig:v2-vs-v25-time}. This allows TabPFN-2.5 to scales inference to datasets with up to $50{,}000$ rows and 2{,}000 features. Furthermore, we found large speed gain in this adoption of FlashAttention-3~\citep{shah2024fa3} and parallel evaluation across multiple GPUs. %

\paragraph{Creating fast, deployable models.}
To improve deployment flexibility, we developed a proprietary distillation engine that, given a training data set, outputs a multi-layer perceptron (TabPFN-2.5-as-MLP) or tree ensemble classifier (TabPFN-2.5-as-TreeEns) whose performance is close to the one of TabPFN on this dataset (see Figure~\ref{fig:mlp-distil-performance}). 
In contrast to TabPFN, this resulting MLP or tree ensemble classifier is dataset-specific, does not perform in-context learning, takes as input a single data point, and has very low latency and memory footprint for making predictions. It can also be seamlessly integrated into existing production pipelines, including those constrained by latency, interpretability, or regulatory requirements that hinder a change in the class of models being deployed.
This increases TabPFN-2.5's practical use in real-world decision systems. Other types of models could easily be supported.

\begin{figure}[htbp]
\centering
\begin{minipage}{0.57\textwidth}  %
\centering
\renewcommand{\arraystretch}{1.05}    %
\setlength{\tabcolsep}{3pt}           %

\small
\begin{tabular}{lccccc}
\toprule
\textbf{Model} & \textbf{Rows} & \textbf{Feat.} & \textbf{Type} & \textbf{Depth} & \textbf{Inference mode} \\
\midrule
TabPFN-v1 & 1{,}000 & 100 & Num. & 8 & ICL \\
TabPFN-v2 & 10{,}000 & 500 & Mixed & 12 & ICL \\
TabPFN-2.5 & 50{,}000 & 2000 & Mixed & 18--24 & ICL+MLP/Trees \\
\bottomrule
\end{tabular}
\captionof{table}{\textbf{Summary of TabPFN model variants.} Max Rows and Features are the recommended maximum sizes. The models also fit larger datasets but are not built and evaluated for these settings.}
\label{tab:model_summary_side}
\end{minipage}%
\hfill
\begin{minipage}{0.39\textwidth}
\centering
\includegraphics[width=1.0\linewidth]{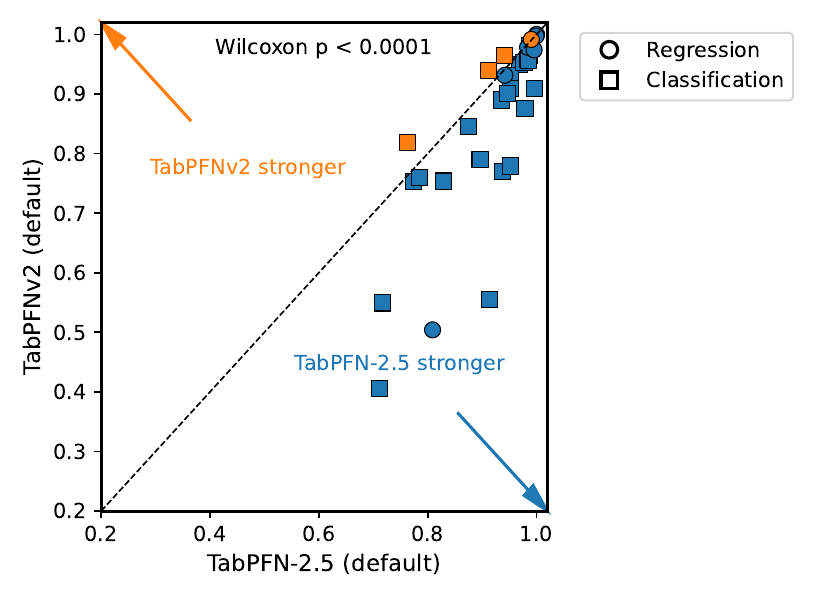}
\captionof{figure}{\textbf{TabPFN-2.5 clearly outperforms TabPFNv2.} We show normalized performance for each dataset of the TabPFNv2 subset of TabArena. TabPFN-2.5 often performs much better 
and is never much worse.}
\label{fig:v2-v2.5_side}
\end{minipage}
\end{figure}

\section{Experimental Results}\label{sec:results} 

We first demonstrate state-of-the-art performance on the industry standard benchmark TabArena and using our own benchmarking framework. Then, we report our advances to reduce inference latency. Finally, we demonstrate that TabPFN-2.5 yields new state-of-the-art performance for causal machine learning.

\subsection{Performance on the Industry Standard Benchmark TabArena}
\label{sec:tabarena-result}

TabArena \citep{erickson2025tabarena} is the most curated tabular benchmark, based on the largest number of candidate datasets considered, and created by open-source contributors from a wide range of institutions. It will appear at the NeurIPS 2025 Datasets \& Benchmarks track and is thus most up-to-date. In particular, it compares a large class of recent models, including tree-based models like CatBoost \citep{prokhorenkova2018catboost}, LightGBM \citep{lightgbm} or XGBoost \citep{chen2016xgboost}, as well as newer deep-learning models like RealMLP \citep{holzmuller2024realmlp}, TabM \citep{gorishniy2024tabm}, ModernNCA \citep{ye2025revisitingnearestneighbortabular} or xRFM \citep{beaglehole2025xrfmaccuratescalableinterpretable}, and other Tabular Foundation Models like TabICL \citep{qu2025tabicl}, TabDPT \cite{ma2025tabdptscalingtabularfoundation}, LimiX \citep{zhang2025limix}, Mitra \citep{zhang2025mitramixedsyntheticpriors} or TabPFNv2 \citep{Hollmann2025tabpfnv2}. We follow the paper's recommendation to benchmark on ``TabArena-Lite'', which is a cheaper but representative version of the full benchmark using only one test fold. The benchmark contains a set of 51 datasets selected from 1053 to be representative of real-world tabular data. See \citet{erickson2025tabarena} for the list of datasets.

\begin{figure}[htb]
\centering
\begin{minipage}[t]{.48\textwidth}
    \centering
    \includegraphics[scale=0.9]{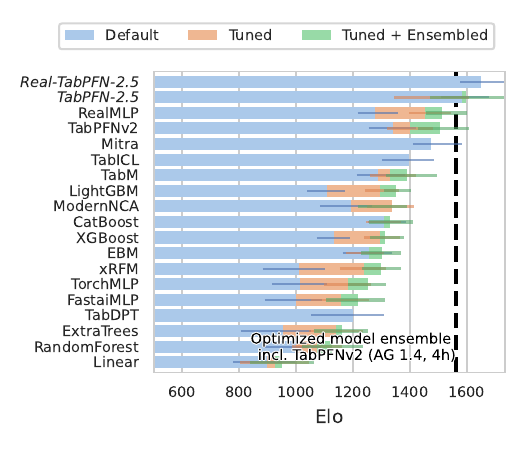}
    \end{minipage} 
\begin{minipage}[t]{.48\textwidth}
    \centering
    \includegraphics[scale=0.9]{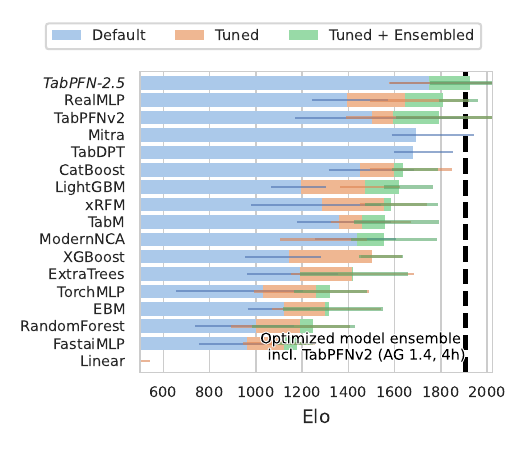}
\end{minipage}
\caption{TabArena-Lite results on \textbf{classification} (left) and \textbf{regression} (right), restricted to datasets with less than \textbf{10K training samples and 500 features}. Note that tuning for TabPFN-2.5 is only based on 60 random configs compared to 200 for the baselines. The vertical dotted line stands for AutoGluon 1.4 extreme mode tuned for 4 hours, an ensemble of models including TabPFNv2 \citep{autogluon_tabular}.}
\label{fig:tabarena-classification-regression-10k-500}
\end{figure}

\paragraph{Pushing the limit on small to medium-sized datasets.} Figure \ref{fig:tabarena-classification-regression-10k-500} shows results for TabPFN-2.5 on TabArena-Lite with up to 10{,}000 data points and 500 features, demonstrating that TabPFN-2.5, in a forward pass, outperforms the wide range of existing tabular prediction methods. On classification, TabPFN-2.5 in a forward pass outperforms AutoGluon 1.4, an ensemble tuned for four hours and including best other methods (even TabPFNv2). Using our Real-TabPFN-2.5 variant fine-tuned on real datasets (deduplicated from TabArena datasets) widens the lead even further. On the other hand, our regression model benefits much more from tuning and outperforms AutoGluon 1.4 after being tuned for 60 configurations.

\begin{figure}[htbp]
\centering
\begin{minipage}[t]{.48\textwidth}
    \centering
    \includegraphics[scale=0.9]{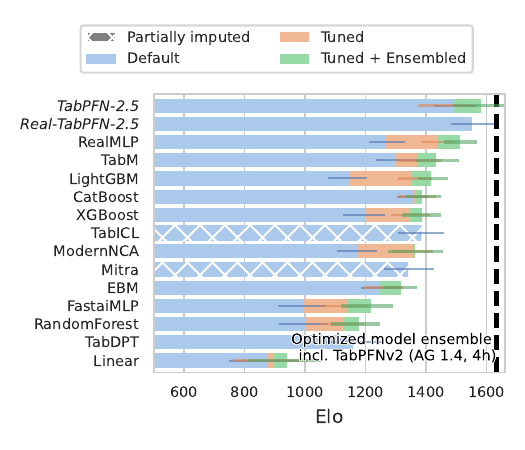}
    \end{minipage} 
\begin{minipage}[t]{.48\textwidth}
    \centering
    \includegraphics[scale=0.9]{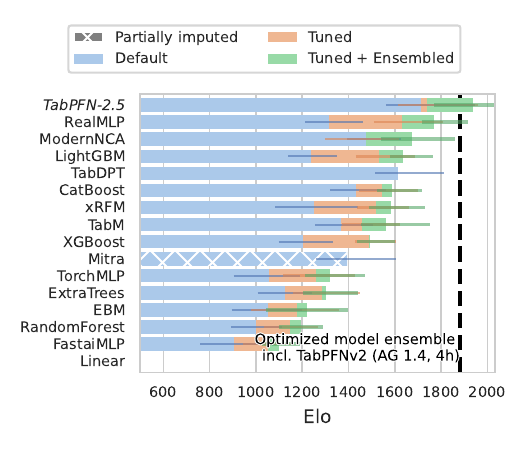}
\end{minipage}
\caption{TabArena-Lite results on \textbf{classification} (left) and \textbf{regression} (right), evaluated on \textbf{all datasets}, going up to \textbf{100K training rows and 2K features}. Note that tuning for TabPFN-2.5 is only based on 60 random configs compared to 200 for the baselines, and that we removed the "dt-pfn" option from our tuning search space for the 4 largest datasets in the benchmark to reduce the tuning time. The vertical dotted line stands for AutoGluon 1.4 extreme mode tuned for 4 hours, an ensemble of models including TabPFNv2 \citep{autogluon_tabular}.}
\label{fig:tabarena-classification-regression}
\end{figure}

\paragraph{Scaling to larger datasets.} Figure \ref{fig:tabarena-classification-regression} shows a similar experiment on all the TabArena datasets, with up to 100{,}000 data points and 2{,}000 features, clearly ranking TabPFN-2.5 as the best default model, and outperforming (for regression datasets) or approaching (for classification datasets) AutoGluon 1.4 (tuned for 4 hours) when tuned. Again, we highlight the very strong default performance of Real-TabPFN-2.5 on these larger classification datasets, beating in one forward pass any other tuned and ensembled model.

\paragraph{A significant improvement upon TabPFNv2.} Comparing the default performance of TabPFN-2.5 and TabPFNv2, we see a big leap in performance in Figure \ref{fig:tabarena-classification-regression-10k-500}. In addition, looking at performance on each dataset in TabArena (TabPFNv2 compatible subset) in Figure \ref{fig:v2-v2.5_side}, we see that TabPFN-2.5 clearly outperforms TabPFNv2 on almost all datasets, and is never much worse. In Appendix \ref{app:tabarena-detailed}, we detail the results on TabArena-Lite, showing the pairwise win rates of the different models, and comparing TabPFN-2.5 to other foundation models like TabICL \citep{qu2025tabicl}, TabDPT \cite{ma2025tabdptscalingtabularfoundation} or LimiX \citep{zhang2025limix}, each time restricting ourselves to the subset of datasets compatible with these models.

\subsection{Performance on Internal Benchmarks}

\paragraph{A diverse internal benchmark.} In addition to the public TabArena benchmark, we built our own benchmarking framework using proprietary data. It includes over 100 use cases from healthcare, finance, insurance, retail and manufacturing. This benchmark focuses on comparing to  gradient-boosted decision tree libraries that are frequently used in industry (XGBoost \citep{chen2016xgboost}, CatBoost \citep{prokhorenkova2018catboost}, LightGBM \citep{lightgbm}), both in their default version and tuned for one hour.
In all cases, we show the results of three standard gradient-boosted tree libraries (LightGBM, XGBoost and CatBoost). We tune all of the baselines for 1hr, using random search on the established search spaces from \cite{Hollmann2025tabpfnv2}. TabPFN is tuned using our AutoTabPFN system, resulting in a tuned and ensembled model.

\begin{centering}
\begin{figure}[htbp]
    \centering
    \begin{subfigure}[t]{0.24\textwidth}
        \centering
        \includegraphics[width=\textwidth]{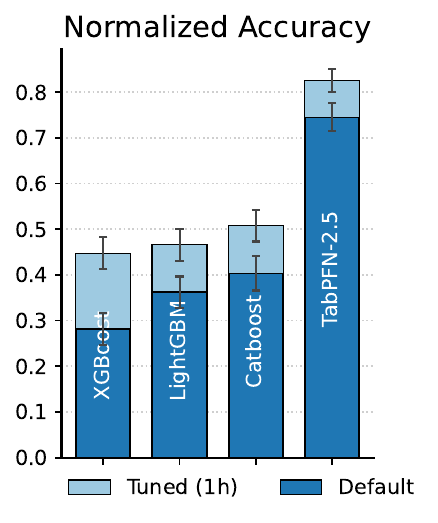}
    \end{subfigure}
    \hspace{-15pt}
    \begin{subfigure}[t]{0.24\textwidth}
        \centering
        \includegraphics[width=\textwidth]{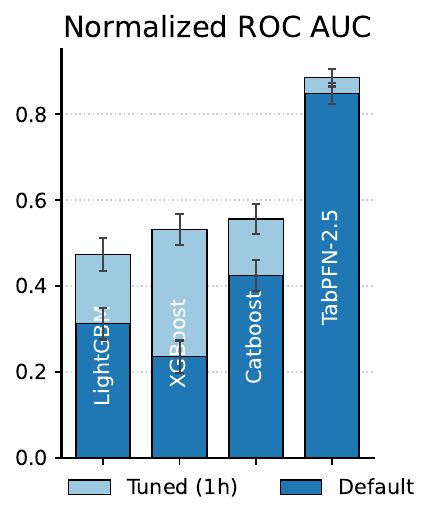}
    \end{subfigure}
    \begin{subfigure}[t]{0.24\textwidth}
        \centering
        \includegraphics[width=\textwidth]{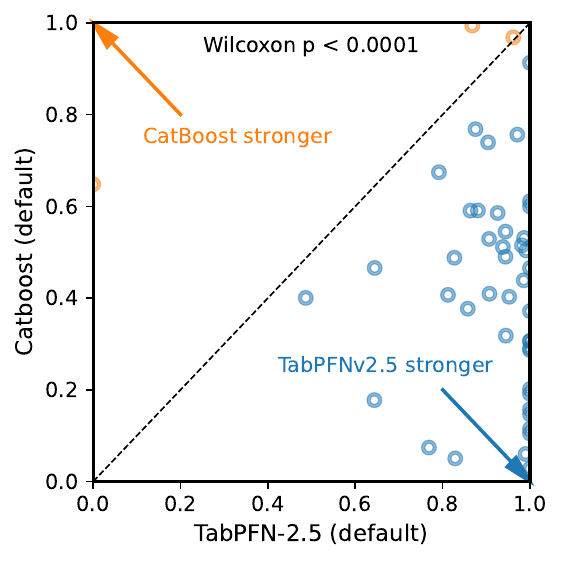}
    \end{subfigure}
    \begin{subfigure}[t]{0.24\textwidth}
        \centering
        \includegraphics[width=\textwidth]{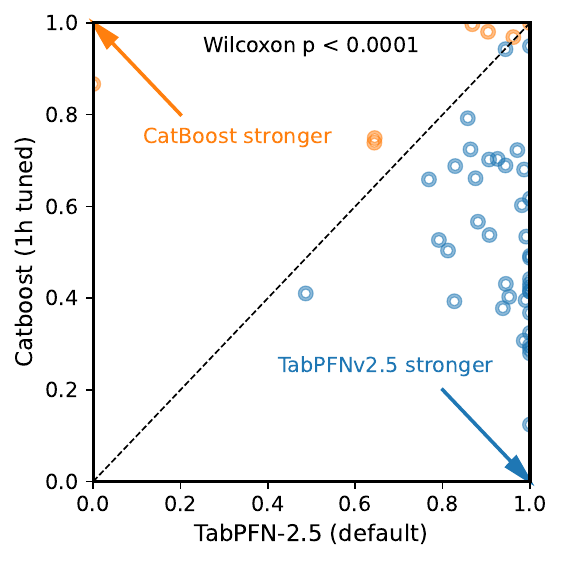}
    \end{subfigure}

    \caption{
        Results from our internal benchmark on \textbf{classification datasets with up to 50,000 data points}. More details on the normalization is available in Appendix \ref{app:additional-internal-results}. In the scatter plots (right), each point represents a different dataset from our internal benchmark, and the axes measure the normalized performance of TabPFN-2.5 and CatBoost (either default or tuned for 1 hour) on this dataset. 
    }
    \label{fig:internal_upto50k_classification}
\end{figure}
\end{centering}

\begin{centering}
\begin{figure}[htbp]    
    \centering
    \begin{subfigure}[t]{0.24\textwidth}
        \centering
        \includegraphics[width=\textwidth]{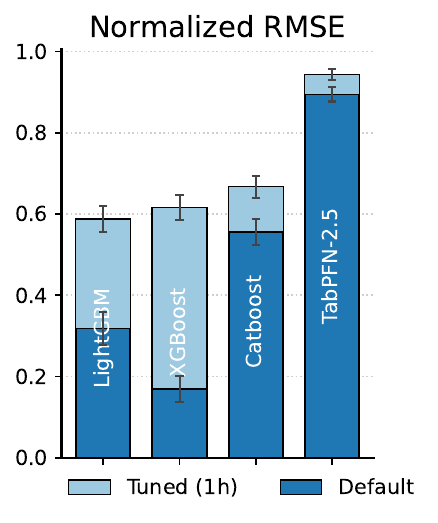}
    \end{subfigure}
    \begin{subfigure}[t]{0.24\textwidth}
        \centering
        \includegraphics[width=\textwidth]{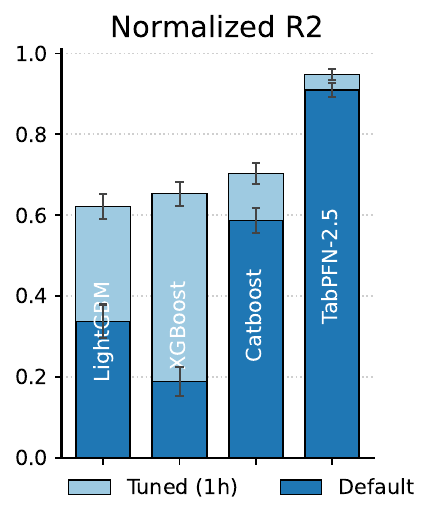}
    \end{subfigure}
    \begin{subfigure}[t]{0.24\textwidth}
        \centering
        \includegraphics[width=\textwidth]{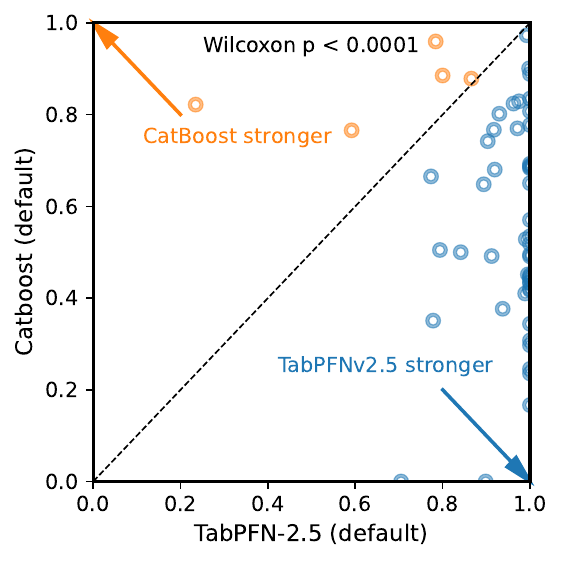}
    \end{subfigure}
    \hfill
    \begin{subfigure}[t]{0.24\textwidth}
        \centering
        \includegraphics[width=\textwidth]{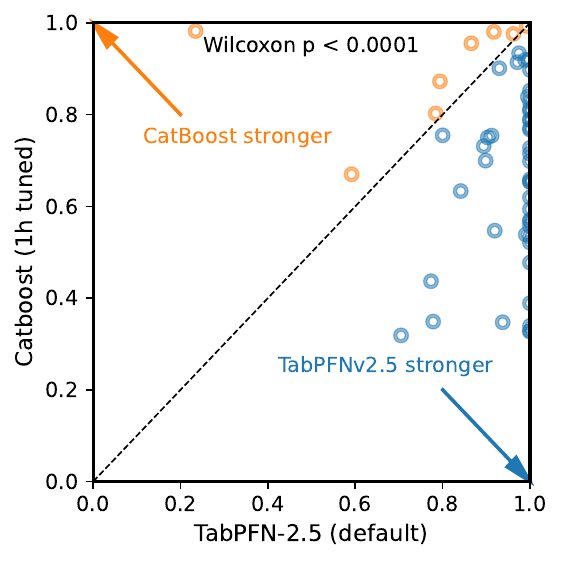}        
    \end{subfigure}
    \caption{
        Results from our internal benchmark on \textbf{regression datasets with up to 50,000 data points}. More details on the normalization is available in Appendix \ref{app:additional-internal-results}.  In the scatter plots (right), each point represent a different dataset from our internal benchmark, and the axis measure the normalized performance of TabPFN-2.5 and CatBoost (either default or tuned for 1 hour) on this dataset
    }
    \label{fig:internal_upto50k_regression}
\end{figure}
\end{centering}

\paragraph{TabPFN-2.5 shows strong results up to 50{,}000 samples and 2{,}000 features.} Figure \ref{fig:internal_upto50k_classification} and Figure \ref{fig:internal_upto50k_regression} show results on our internal benchmark for classification and regression datasets with up to 50,000 data points and 500 features. We can see on these figures that TabPFN outperforms in one forward pass all our tuned baselines. In Section \ref{app:additional-internal-results}, we also show strong results on datasets with 500 to 2,000 features, and provide more details on how we normalize the performance of each model across datasets.

\begin{figure}
    \centering
    \includegraphics[width=0.7\textwidth]{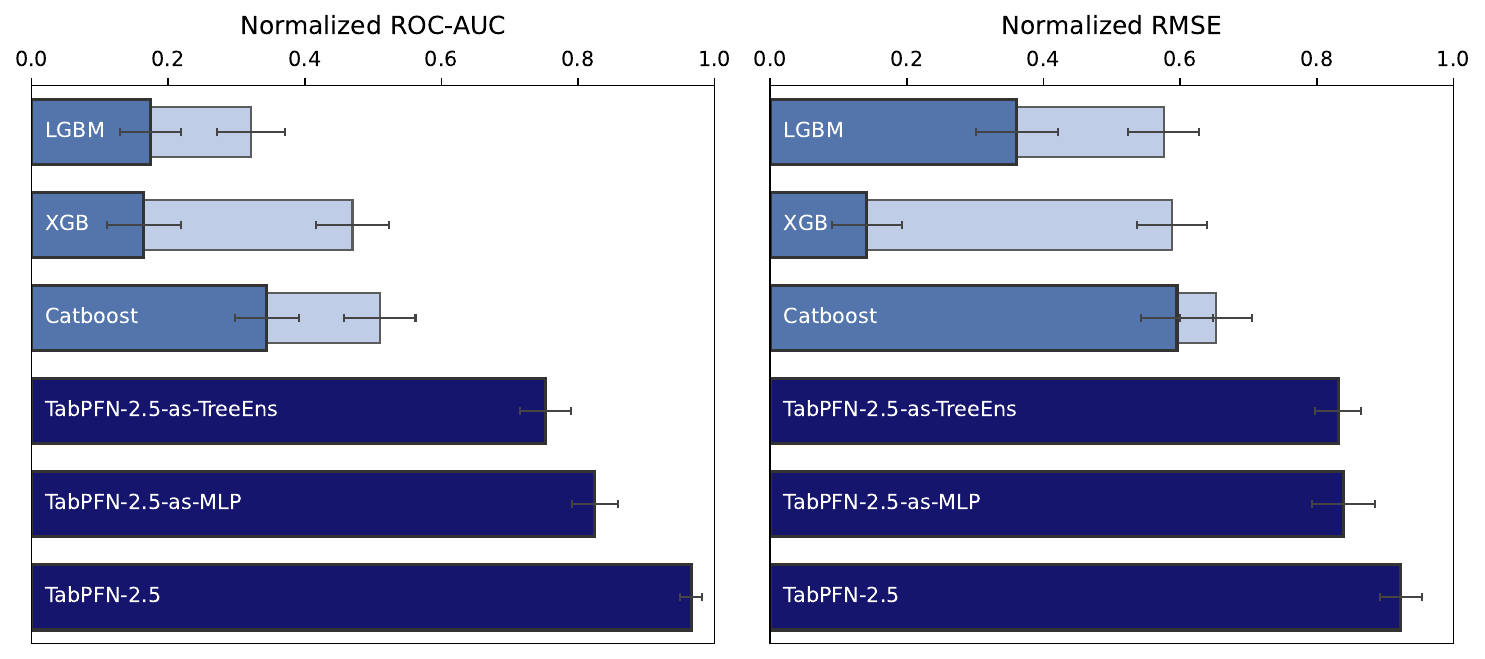}
    \caption{\textbf{TabPFN-as-MLP and  TabPFN-as-TreeEns still outperform tree-based models} while having much faster inference speed than TabPFN. For baseline, light blue represents performance when tuned for 1 hour, and darker blue default performance. For TabPFN, we report default performance. All scores are on from our internal benchmark, up to 10K data points.}
    \label{fig:mlp-distil-performance}
\end{figure}

\subsection{Measuring TabPFN-2.5 Training and Inference Speed}
\label{sec:inference-latency}
\begin{figure}[tb]
    \centering
    \begin{subfigure}[t]{0.49\textwidth}
        \centering
        \includegraphics{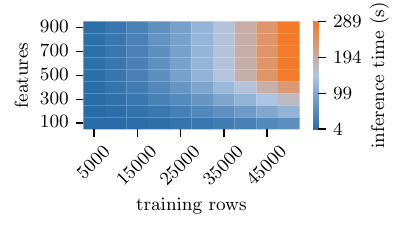}
        \caption{one H100 GPU}
    \end{subfigure}
    \hfill
    \begin{subfigure}[t]{0.49\textwidth}
        \centering
        \includegraphics{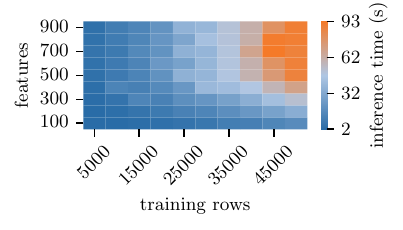}
        \caption{four H100 GPUs}
    \end{subfigure}
    \caption{
        Time taken, in seconds, to fit TabPFN-2.5 classification models on various training set sizes, and then make predictions on $500$ test rows.
        Figure~\ref{fig:inference-time-by-gpu} in Section~\ref{sec:speed-common-gpus} reports results for regression, alongside performance on A100 and T4 GPUs.
    }
    \label{fig:inference-time-heatmap}
\end{figure}

Figure~\ref{fig:inference-time-heatmap} shows how TabPFN-2.5 classification speed scales with training set size, when using one or four GPUs, as we vary the number of rows and columns in the dataset. The time measured includes both the time to process the training rows (equivalent to the combination of ``training'' a classical ML model) and ``prediction'' time on test rows.
We can observe the expected scaling in $\mathcal{O}(r^2 \min(c, 500) + r\min(c, 500)^2)$, where $r$ is the number of rows and $c$ is the number of columns, due to dual attention over rows and capped per-estimator feature subsampling at 500 features.
Section~\ref{sec:speed-common-gpus} contains results for regression, performance on common models of GPU, for reference, and a measurement of the speedup from TabPFNv2. The inference speed reported here reflects the latency of the full in-context learning model.

\subsection{Fast Inference with our distillation engine}
\label{sec:tabpfn-as-mlp}

We benchmark our distillation engine which, given a training data set, outputs a multi-layer perceptron (TabPFN-2.5-as-MLP) or tree ensemble classifier (TabPFN-2.5-as-TreeEns), against tuned LightGBM, XGBoost, and CatBoost models, as well as the standard TabPFN-2.5 model, on our curated collection of internal open source datasets with less than 10k data points. Figure \ref{fig:mlp-distil-performance} illustrates representative test-split performance. Empirically, TabPFN-2.5-as-MLP and  TabPFN-2.5-as-TreeEns offer competitive accuracy while reducing inference cost, making it attractive for high-throughput or resource-constrained deployment scenarios.

\subsection{TabPFN for Causal Inference}

\begin{figure}
\centering
\begin{minipage}[t]{.48\textwidth}
    \centering
    \includegraphics[scale=0.475]{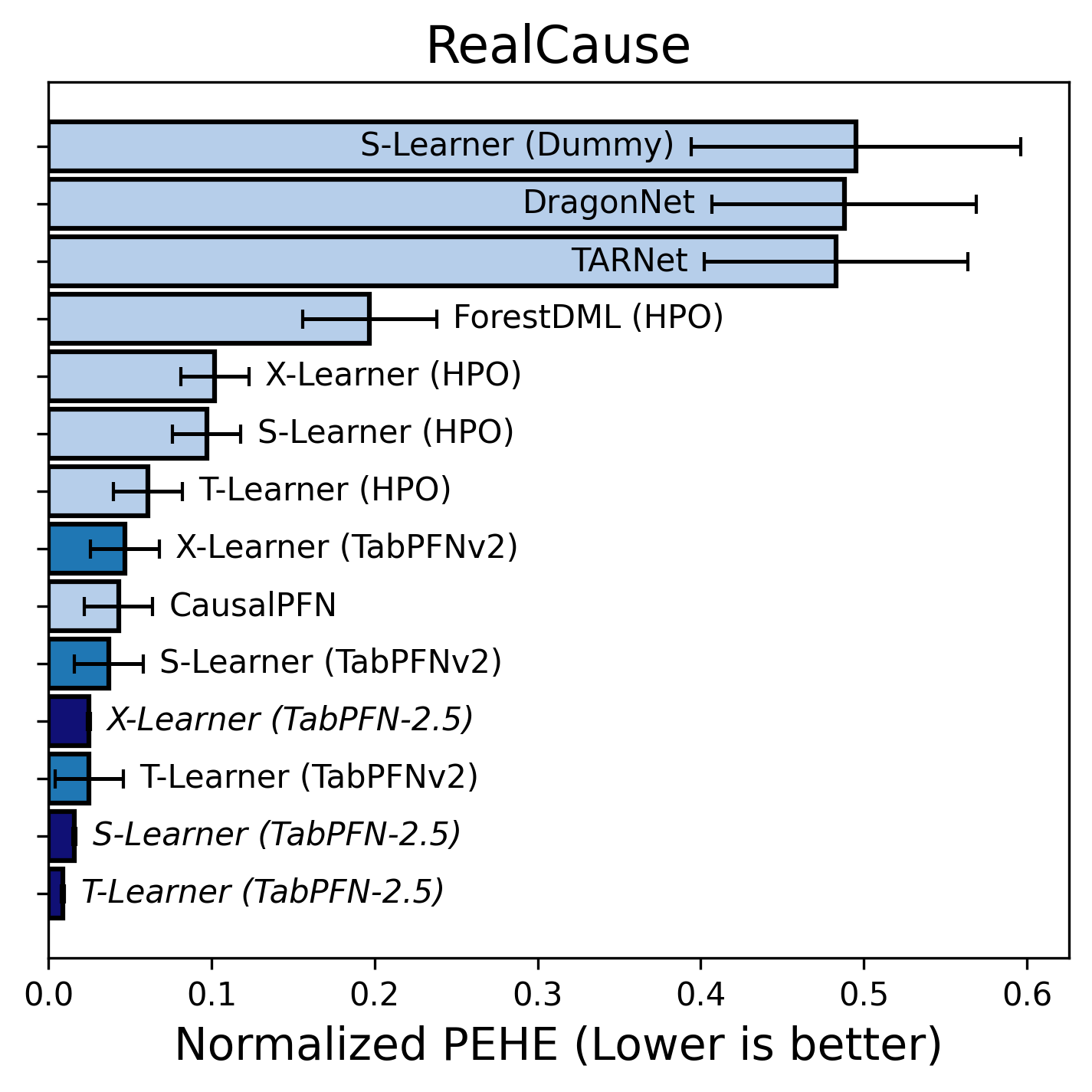}
    \caption{
        PFN-based CATE estimators dominate RealCause, outperforming specialized tree- and deep-learning-based methods for causal inference. Choice of propensity and outcome model is important for CATE estimation. 
    }
    \label{fig:realcause}
    \end{minipage} \hfill
\begin{minipage}[t]{.48\textwidth}
    \centering
    \includegraphics[scale=0.475]{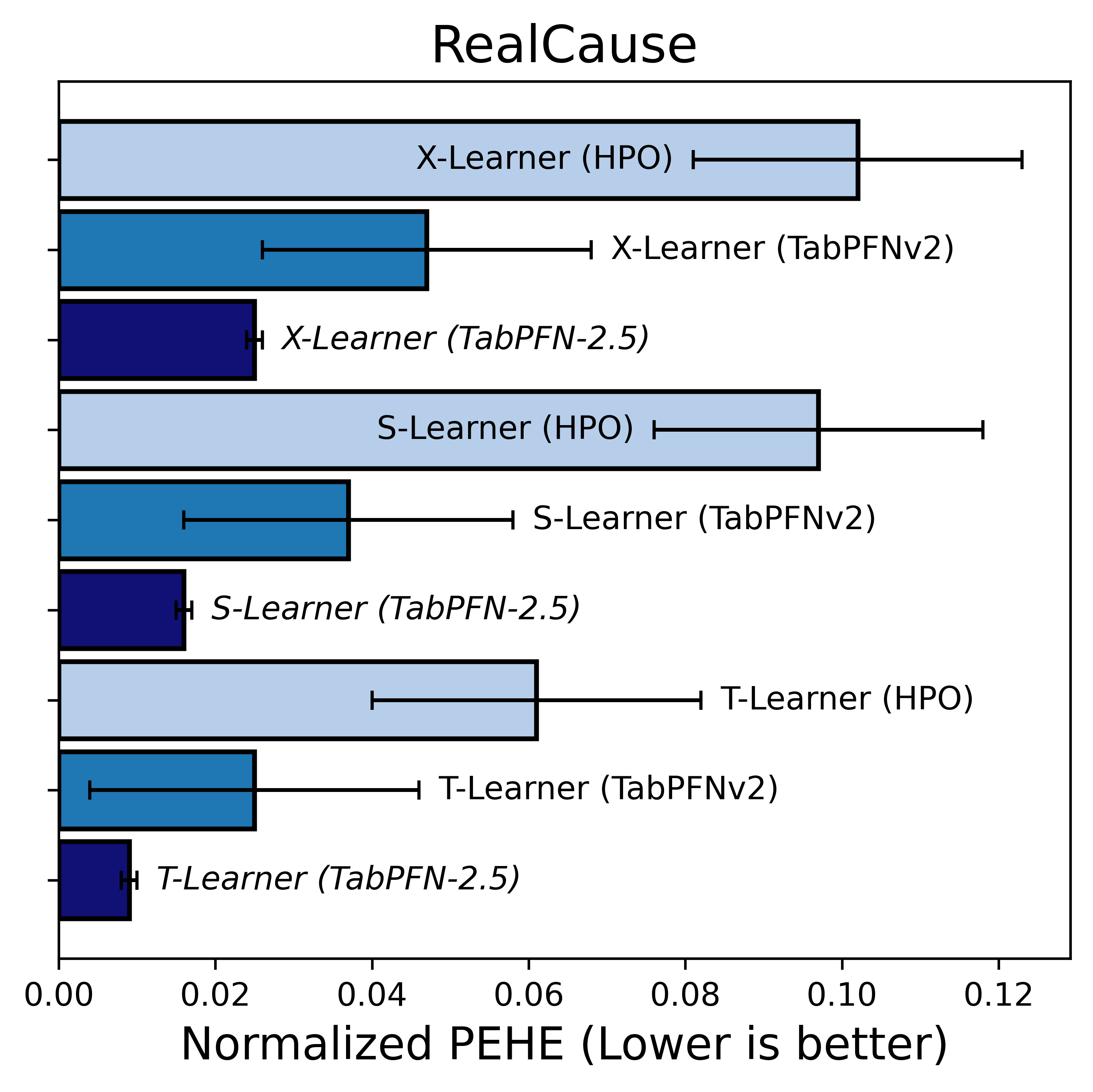}
    \caption{
        Improvements in base model predictive performance transfer to improved performance in CATE estimation. Our new model, TabPFN-2.5, is the strongest choice of base model for all meta-learners.
    }
    \label{fig:realcause_tabpfn}
\end{minipage}
\end{figure}

\paragraph{RealCause Benchmark.} To systematically evaluate TabPFN’s potential as a causal estimator, we leverage the RealCause benchmark \cite{neal_realcause}, a semi-synthetic benchmark which begins with real-world randomized control trial (RCT) data and synthetically creates observable confounding effects.\footnote{Descriptions of the ACIC-2016, IHDP, and Lalonde-PSID and Lalonde-CPS datasets are provided in Appendix Table \ref{tab:realcause_data}.} We measure the Precision in Estimating Heterogeneous Effects (PEHE), which corresponds to the root-mean-squared error between predicted and RealCause's ground-truth CATE values\footnote{For a description of the CATE estimation task and common estimators, please refer to Appendix \ref{sec:causal_inference}.}. In Figure \ref{fig:realcause}, we show that PFN-based methods for CATE-estimation dominate the leaderboard, occupying the first seven positions. TabPFN-2.5 applied as a T-Learner, a simple two-model approach that fits a separate model to the treatment and control observations, achieves the strongest overall performance, outperforming specialized tree- and deep-learning-based methods \cite{wager_causal_forest}. We also observe in Figure \ref{fig:realcause_tabpfn} that for each of our three meta-learners, TabPFN-2.5 performs better out-of-the-box than TabPFNv2 and HPO\footnote{Hyperparameter optimization is run for 60 seconds on an H100 per propensity and outcome model using FLAML \cite{wang_flaml}.}. This result shows that improvements in base model predictive performance transfer to the problem of causal inference.

\paragraph{Foundation Models for Causal Inference.} While we show strong results in unconfounded settings, real-world causal inference often involves imperfect data and latent confounders. A growing line of work aims to pre-train PFNs explicitly for causal reasoning—for example, predicting interventional outcomes or learning causal structures directly \citep{balazadeh_causalpfn, dhir_interv, feuerriegel_causalfm, robertson_dopfn, sauter_activa}. We view this as one of the most exciting frontiers for foundation models: extending TabPFN’s reasoning from predicting \textit{what is} to inferring \textit{what would happen if}, and ultimately, \textit{understanding why}.

\section{How to Get Optimal Fit + Predict Speed from TabPFN-2.5}
\label{sec:speed-common-gpus}
To achieve good performance, we recommend the following:
\begin{itemize}
    \item \textbf{Use a dedicated GPU or GPUs}: We recommend NVIDIA H100 or A100 GPUs. Any dedicated GPU supported by PyTorch is compatible, but some models may not have enough memory for larger datasets or perform slowly.
    Integrated GPUs, MPS (Apple Silicon), and CPUs are also supported, but are only suitable for small datasets.
    \item \textbf{Use multiple GPUs}: For larger datasets, fit + predict time can be dramatically reduced by parallelizing inference over several GPUs. To enable this, set the \texttt{device} parameter of \texttt{TabPFNClassifier} and \texttt{TabPFNRegressor}.
    \item \textbf{Use batch inference}: Unless the fitted-model cache is enabled (see below), the model is retrained each time \texttt{.predict()} is called. This means that it is much faster to make a prediction for all your test points in a single \texttt{.predict()} call. If you run out of memory, split the test points into batches of $1000$ to $10000$ and call \texttt{.predict()} for each batch.
    \item \textbf{Use PyTorch 2.8 or above}: TabPFN-2.5 also supports earlier versions of PyTorch, but these may have lower performance.
    \item \textbf{For small datasets, enable the fitted-model cache}:
        This is an experimental feature that trains and stores the model during \texttt{.fit()}, making subsequent \texttt{.predict()} calls fast by using a KV-Cache. It is enabled by setting the \texttt{fit\_mode} parameter of \texttt{TabPFNClassifier} and \texttt{TabPFNRegressor} to \texttt{fit\_with\_cache}.
        However, with this setting classification models will consume approximately 6.1 KB of GPU memory and 48.8 KB of CPU memory per cell in the training dataset (regression models about $25$\% less), thus it is currently only suitable for small training datasets. For larger datasets and CPU-based inference, we recommend the TabPFN-as-MLP/TreeEns.
    \item If speed is important for your application, you may consider optimizing the \texttt{memory\_saving\_mode} and \texttt{n\_preprocessing\_jobs} parameters of \texttt{TabPFNClassifier} and \texttt{TabPFNRegressor}. See the code documentation for further information.
\end{itemize}

Figure~\ref{fig:inference-time-by-gpu} in the appendix shows the inference latency you can expect for three common models of GPU, when using one or four GPUs.
It also shows the maximum dataset size that fits in memory for each GPU.

\section{License and Availability}\label{sec:license}

We release TabPFN-2.5 under our \texttt{TABPFN-2.5 License v1.0} designed to be permissive for research and internal evaluation. It \textit{explicitly allows} testing, evaluation, and internal benchmarking, so an organization can download the model and run preliminary assessments on its own datasets.

The key restriction is that the model, its derivatives, and its outputs cannot be used for any commercial or production purpose. This includes, but is not limited to, revenue-generating products, competitive benchmarking for procurement, client deliverables, or using the model's results for internal commercial decision-making.

For production use cases, we offer a \textit{Commercial Enterprise License}. This provides access to our proprietary high-speed inference engine, dedicated support, integration tooling, and other internal models.

Please contact us at sales@priorlabs.ai for commercial licensing inquiries. The full non-commercial mode license text can be found at \url{https://huggingface.co/Prior-Labs/tabpfn_2_5/blob/main/LICENSE}.

\subsection{Cloud API}
We provide a managed TabPFN-2.5 cloud endpoint, which runs on our optimized GPU infrastructure. This is the recommended option for users who do not have a dedicated local GPU or for those who wish to use TabPFN commercially without purchasing a full on-premise license.

The API is accessible via a simple Python SDK\footnote{The Python client SDK is available on PyPI: \url{https://github.com/PriorLabs/TabPFN-client}} (\texttt{pip install tabpfn-client}) or a standard REST API, allowing for integration into both non-commercial and commercial applications.

\section{Conclusion
and The Road Ahead}\label{sec:conclusions}

We are excited about this release.
Taken together, our experiments on public (TabArena) and private benchmarks demonstrate that TabPFN-2.5 sets a new state-of-the-art for tuning-free tabular models. %
Built for datasets up to 50,000 rows and 2,000 features, TabPFN-2.5 matches the performance of complex 4-hour-tuned ensembles - ensembles that even include our previous TabPFNv2 - and in a forward pass outperforms  any other tuned model on the unrestricted public TabArena benchmark (which contains datasets with up to 100{,}000 training data points).

The next step is scaling to datasets with millions of rows. We are actively developing new techniques - including retrieval, fine-tuning, and novel architectures - and anticipate that systems based on Tabular Foundation Models (TFMs) will define state-of-the-art performance for datasets with millions of data points very soon.

Our broader vision beyond this release is to tackle the entire stack of problems with tabular-like data, including time series, multimodal tabular data, causal inference, unsupervised tasks, integration of domain knowledge and decision support, ultimately building the core intelligence engine for reasoning over structured and multimodal data.

\FloatBarrier

\bibliographystyle{unsrtnat}
\bibliography{bib}

\newpage
\appendix

\section{Contributors}\label{app:contributors}

\subsection*{Model dev \& Deployment}

L\'eo Grinsztajn,
Klemens Flöge,
Oscar Key,
Felix Birkel,
Brendan Roof,
Phil Jund,
Benjamin Jäger,
Adrian Hayler,
Dominik Safaric,
Simone Alessi,
Felix Jablonski,
Mihir Manium,
Rosen Yu,
Anurag Garg,
Jake Robertson,
Shi Bin (Liam) Hoo,
Vladyslav Moroshan,
Magnus Bühler,
Lennart Purucker,
Bernhard Schölkopf,
Noah Hollmann,
Frank Hutter

\vspace{0.5em}

\subsection*{Distribution \& Product}

Clara Cornu,
Lilly Charlotte Wehrhahn,
Alessandro Bonetto,
Sauraj Gambhir
\vspace{0.75em}
\\
\textit{Special thanks to Samuel Müller for helpful discussions and support.}

\vspace{1em}

\section{TabPFN Use Case Overview}\label{app:use_cases}

TabPFNv2 has been applied to a broad set of use cases. We now list 100 published use cases across different industries.

\section*{Healthcare and Life Sciences}

We collected 51 published TabPFN use cases in this area, by far more than in any other area; we attribute this partly to the scarcity of data in healthcare and life sciences, and partly to the open publishing culture in this area. Use cases span oncology, neurology, cardiology, psychiatry, nephrology, and pharmacology. Applications include diagnosis, prognosis, and treatment response prediction from multimodal clinical, imaging, and omics data, often under severe data scarcity.

\begin{enumerate}

\item TabPFN was applied to distinguish cancer patients from healthy individuals using immune system profiles from peripheral blood, facilitating predictions of immunotherapy responses \cite{hc_usecase1_bostongene_tabpfn}. \href{https://www.linkedin.com/pulse/how-bostongene-utilized-tabpfn-identify-immune-system-profiles-vexle/}{Link}

\item A machine learning model employing TabPFN was developed for non-invasive diagnostic prediction of minimal change disease in patients with nephrotic syndrome, utilizing clinical biomarkers \cite{hc_usecase2_mcd_scirep}. \href{https://www.nature.com/articles/s41598-024-73898-4}{Link}

\item TabPFN was integrated into a system for analyzing T-cell receptor repertoires combined with clinical biomarkers to forecast immunotherapy outcomes in cancer patients, as explored by researchers at BostonGene \cite{hc_usecase3_immunotypes_cancercell}. \href{https://www.cell.com/cancer-cell/fulltext/S1535-6108(24)00132-6}{Link}

\item TabPFN enabled early detection of stillbirth risks through analysis of cardiotocography data, supporting improved prenatal care \cite{hc_usecase4_stillbirth_slas}. \href{https://www.sciencedirect.com/science/article/pii/S2472630324000852}{Link}

\item Predictive modeling for postoperative outcomes following anterior cervical corpectomy utilized TabPFN to assess patient demographics and surgical parameters \cite{hc_usecase5_acc_asj}. \href{https://pmc.ncbi.nlm.nih.gov/articles/PMC11366553/}{Link}

\item A hybrid model incorporating TabPFN was introduced to predict dementia progression in Parkinson's disease patients, handling small datasets and missing values effectively \cite{hc_usecase6_pd_dementia_lightgbm_tabpfn}. \href{https://journals.sagepub.com/doi/full/10.1177/20552076241272585}{Link}

\item A machine learning model based on TabPFN was developed to predict 90-day unfavorable outcomes in stroke patients with distal vessel occlusions using CT perfusion imaging \cite{hc_usecase7_dmvo_ajnr}. \href{https://www.ajnr.org/content/early/2024/10/28/ajnr.A8547.abstract}{Link}

\item TabPFN was utilized in chemoproteomics for identifying small-molecule fragment-protein interactions, aiding ligand discovery in drug development \cite{hc_usecase8_chemoproteomics_science}. \href{https://www.science.org/doi/abs/10.1126/science.adk5864}{Link}

\item TabPFN facilitated the prediction of non-invasive ventilation outcomes in patients with acute hypoxemic respiratory failure, supporting early identification of treatment failures \cite{hc_usecase9_niv_tabpfn}. \href{https://www.researchgate.net/profile/Antonio-Esquinas/publication/393595503_Early-prediction-of-non-invasive_ventilation_outcome_using_the_TabPFN_machine_learning_model_a_multi-centre_validation_study/links/68718bc56e247f362b18c4b8/Early-prediction-of-non-invasive-ventilation-outcome-using-the-TabPFN-machine-learning-model-a-multi-centre-validation-study.pdf}{Link}

\item An interpretable Transformer-based model leveraging TabPFN was created to predict intravenous immunoglobulin resistance in pediatric patients with Kawasaki disease \cite{hc_usecase10_kawasaki_tabpfnv2}. \href{https://journals.plos.org/plosone/article?id=10.1371/journal.pone.0327564}{Link}

\item TabPFN was employed in visual representation techniques for prostate cancer diagnosis, converting clinical biomarkers and symptom data into formats suitable for analysis \cite{hc_usecase51_prostate_visual_rep}. \href{https://www.mdpi.com/2306-5354/11/7/635}{Link}

\item TabPFN was used to combine clinical, MR morphological, and delta-radiomics features to predict lymphovascular invasion in invasive breast cancer patients \cite{hc_usecase11_lvi_breast_tabpfn}. \href{https://journals.sagepub.com/doi/full/10.1177/15330338251362050}{Link}

\item TabPFN is proposed to predict mental health trajectories through digital phenotyping, enabling proactive and personalized interventions in precision psychiatry \cite{hc_usecase12_precision_psychiatry_tabpfn}. \href{https://onlinelibrary.wiley.com/doi/epdf/10.1002/mdr2.70017}{Link}

\item TabPFN contributed to cardiovascular disease risk stratification using clinical features from a large patient cohort, incorporating interpretability techniques \cite{hc_usecase13_ml_health_tabpfn}. \href{https://github.com/Bruno-LSo/ML-Health-TABPFN}{Link}

\item TabPFN outperformed traditional machine learning models for early prediction of acute kidney injury in hospitalized patients, demonstrating generalizability across datasets \cite{hc_usecase14_aki_ssrn_tabpfn}. \href{https://papers.ssrn.com/sol3/papers.cfm?abstract_id=5397006}{Link}

\item TabPFN was integrated into a framework for predicting postoperative mobility and discharge destinations in older adults using sensor data \cite{hc_usecase15_postop_mobility_sensors}. \href{https://www.mdpi.com/1424-8220/25/16/5021}{Link}

\item TabPFN supported the prediction of infant temperament from maternal mental health data, aiding early identification of at-risk infants \cite{hc_usecase16_infant_temperament_tabpfn}. \href{https://www.frontiersin.org/journals/public-health/articles/10.3389/fpubh.2025.1659987/abstract}{Link}

\item TabPFN was employed to characterize clinical risk profiles for complications in type 2 diabetes mellitus patients, focusing on neuropathy and retinopathy \cite{hc_usecase17_t2dm_complications_tabpfn}. \href{https://www.frontiersin.org/journals/endocrinology/articles/10.3389/fendo.2025.1657366/abstract}{Link}

\item TabPFN was extended with a longitudinal-to-cross-sectional transformation to forecast Alzheimer's disease progression on neuroimaging datasets \cite{hc_usecase18_ad_l2c_tabpfn}. \href{https://arxiv.org/abs/2508.17649}{Link}

\item TabPFN supported uncertainty calibration evaluation in medical data using variational techniques \cite{hc_usecase19_uncertainty_vbll_tabpfn}. \href{https://arxiv.org/abs/2509.10048}{Link}

\item TabPFN was applied to predict tumor response to chemotherapy in cholangiocarcinoma patients using RNA expression landscapes \cite{hc_usecase20_cholangio_aacr_tabpfn}. \href{https://aacrjournals.org/clincancerres/article/31/13_Supplement/A020/763312}{Link}

\item TabPFN was incorporated into a generative model framework for tasks like data augmentation and imputation in biomedicine \cite{hc_usecase21_tabpfgen}. \href{https://arxiv.org/abs/2406.05216}{Link}

\item TabPFN facilitated the prediction of gallstone malignancy risks through analysis of associated disease factors \cite{hc_usecase22_gallstone_malignancy_tabpfn}. \href{https://www.mdpi.com/2077-0383/14/17/6091}{Link}

\item TabPFN was used in classifying tuberculosis treatment outcomes based on clinical and sociodemographic data from national registries \cite{hc_usecase23_tb_outcomes_tabpfn}. \href{https://www.researchsquare.com/article/rs-7502054/v1}{Link}

\item TabPFN contributed to early prediction of gestational diabetes using cell-free DNA and genetic scores from early pregnancy blood samples \cite{hc_usecase24_gdm_cfdna_tabpfn}. \href{https://www.medrxiv.org/content/10.1101/2025.09.03.25334985v1}{Link}

\item TabPFN was used for predicting schizophrenia based on sense of agency features, emphasizing interpretability \cite{hc_usecase25_schizophrenia_soa_tabpfn}. \href{https://www.sciencedirect.com/science/article/abs/pii/S187620182500317X}{Link}

\item TabPFN was integrated into a physiologically based pharmacokinetic model for predicting dissolution and absorption of amorphous solid dispersions in drug development \cite{hc_usecase26_pbpk_asd_tabpfn}. \href{https://doi.org/10.1016/j.jconrel.2025.114123}{Link}

\item TabPFN enabled classification of respiratory diseases from sound data, addressing clinical spectrum diversity \cite{hc_usecase27_respiratory_sounds_tabpfn}. \href{https://papers.ssrn.com/sol3/papers.cfm?abstract_id=5529540}{Link}

\item TabPFN was applied to small-data tabular learning in drug discovery, handling data scarcity and distribution shifts \cite{hc_usecase28_drug_discovery_small_data_tabpfn}. \href{https://chemrxiv.org/engage/chemrxiv/article-details/68d29b1cf2aff1677025b18f}{Link}

\item TabPFN facilitated prediction of coronary heart disease risk in patients with cardiovascular-kidney-metabolic syndrome, optimizing evaluation in small samples \cite{hc_usecase29_ckm_chd_tabpfn}. \href{https://pmc.ncbi.nlm.nih.gov/articles/PMC12437168/}{Link}

\item TabPFN was used to predict success of allogeneic stem cell mobilization in donors, aiding transplant therapies \cite{hc_usecase30_stem_cell_mobilization_tabpfn}. \href{https://www.biorxiv.org/content/10.1101/2025.09.17.676674v1.full}{Link}

\item TabPFN contributed to predicting manual strength using anthropometric data, focusing on accuracy and interpretability \cite{hc_usecase31}. \href{https://pubmed.ncbi.nlm.nih.gov/41021732/}{Link}

\item TabPFN supported uncertainty-guided model selection for biomolecule efficacy prediction, enhancing ensemble optimization in drug discovery, as studied at GSK \cite{hc_usecase32}. \href{https://www.arxiv.org/abs/2510.02476}{Link}

\item TabPFN was utilized in a multitask deep learning framework for optimizing in vitro fertilization decisions, including embryo transfer and pregnancy prediction \cite{hc_usecase33}. \href{https://dspace.mit.edu/bitstream/handle/1721.1/162969/zheng-zhengr-meng-eecs-2025-thesis.pdf?sequence=1&isAllowed=y}{Link} 

\item TabPFN enabled a framework for early Long COVID detection through causal gene identification and interpretability \cite{hc_usecase34}. \href{https://www.medrxiv.org/content/10.1101/2025.10.02.25337138v1.full.pdf}{Link}

\item TabPFN was used in a foundation model approach for neoadjuvant therapy recommendations in breast cancer, integrating multi-omics data \cite{hc_usecase35}. \href{https://www.medrxiv.org/content/10.1101/2025.10.03.25337255v1}{Link} 

\item Recent work has demonstrated explainable machine learning pipelines for coronary artery disease stratification from routine clinical data \cite{healthcare_explainable_cad_algorithms2025}. \href{https://www.mdpi.com/1999-4893/18/11/693}{Link}

\item TabPFN facilitated prediction of recurrence and progression in oral potentially malignant disorder patients post-surgery \cite{hc_usecase36}. \href{https://journals.lww.com/international-journal-of-surgery/abstract/9900/artificial_intelligence_for_predicting.3354.aspx}{Link} 

\item TabPFN supported prediction of occult lymph node metastasis in non-small cell lung cancer patients treated with stereotactic ablative radiotherapy \cite{hc_usecase37}. \href{https://www.redjournal.org/article/S0360-3016(25)05890-0/fulltext}{Link}

\item TabPFN was used in stroke diagnosis, addressing dataset imbalance and model interpretability for clinical decisions \cite{hc_usecase38}. \href{https://www.ijsab.com/jsr-volume-9-issue-1/8205}{Link}

\item TabPFN was integrated into a multimodal thesis framework for clinical predictions using tabular and phenotypic data from large-scale projects \cite{hc_usecase39}. \href{https://irep.mbzuai.ac.ae/items/3e3d4c0d-dbcb-4d5b-a23e-e28aea840660}{Link}

\item TabPFN was used to predict diabetes-related hypo- and hyperglycemia during hemodialysis using continuous glucose monitoring data, facilitating improved patient management \cite{hc_usecase40}. \href{https://www.medrxiv.org/content/10.1101/2025.10.24.25338707v1}{Link}

\item TabPFN was applied to enhance diagnosis of hypervascular thyroid nodules using multimodal ultrasound features \cite{hc_usecase52_thyroid_hypervascular_multimodal}. \href{https://pmc.ncbi.nlm.nih.gov/articles/PMC12432950/}{Link}

\item TabPFN was integrated with radiomics and clinical features to predict endovascular treatment success in femoropopliteal chronic total occlusions, supporting interventional planning \cite{hc_usecase53_fp_cto_radiomics}. \href{https://www.researchgate.net/publication/396892115_Radiomics_enhance_the_prediction_of_endovascular_treatment_success_for_femoropopliteal_chronic_total_occlusions_a_proof-of-concept_study}{Link}

\item TabPFN was applied to CorvisST biomechanical indices to classify corneal disorders, improving diagnostic accuracy in ophthalmology \cite{hc_usecase41_corvisst_corneal}. \href{https://pubmed.ncbi.nlm.nih.gov/41130662/}{Link}

\item TabPFN was incorporated into a non-invasive sleep staging framework using respiratory sound features, advancing passive sleep monitoring \cite{hc_usecase42_sleepstage_resp_sounds}. \href{https://www.mdpi.com/1424-8220/25/20/6282}{Link}

\item TabPFN supported prediction of vancomycin blood concentrations to optimize antimicrobial dosing strategies in clinical practice \cite{hc_usecase43_vancomycin_mimic4}. \href{https://journal.china-pharmacy.com/en/article/doi/10.6039/j.issn.1001-0408.2025.19.16/}{Link} 

\item TabPFN was used to predict negative self-rated oral health in adults, identifying risk factors for targeted public-health interventions \cite{hc_usecase44_sroh_jdent}. \href{https://www.sciencedirect.com/science/article/pii/S0300571225006104}{Link} 

\item TabPFN was extended to very high-dimensional feature spaces to enable robust analysis of biomedical data, improving stability and interpretability in clinical applications \cite{hc_usecase45_tabpfn_wide}. \href{https://arxiv.org/abs/2510.06162}{Link}

\item TabPFN predicted gastrointestinal bleeding risk in pediatric Henoch–Schönlein purpura patients, supporting early clinical intervention \cite{hc_usecase50_gibleed_hsp}. \href{https://www.frontiersin.org/journals/physiology/articles/10.3389/fphys.2025.1630807/full}{Link}

\item TabPFN was used as the pre-trained backbone (embeddings + in-context learning) for silica nanoparticle cellular toxicity prediction \cite{other_usecase22_silica_np_toxicity_tabpfn}. \href{https://www.researchsquare.com/article/rs-7735307/v1}{Link}

\end{enumerate}

\section*{Financial Services, Banking, and Insurance}

While we have seen strong customer interest in this area, this is not reflected by the relatively few published use cases (only 3) we managed to collect; we attribute this to the domain's competitive nature and disinclination to publish.

\begin{enumerate}

\item TabPFN was applied to usage-based premium calculations in actuarial science, leveraging driving behavior data from IoT devices \cite{fin_usecase1_nonlife_transformers_actuarial}. \href{https://idp.springer.com/authorize/casa?redirect_uri=https://link.springer.com/article/10.1007/s13385-024-00388-2&casa_token=5LdKiRIXfwEAAAAA:45MEDhjSq66DEqh96gk0NTWrhozhvBbd73mH-oMMuukD0EeHxH1fx3DTtp7h_l04IAjDuJXnpO2uHaHxjw}{Link}

\item TabPFN facilitated cross-selling of health insurance products through deep learning analysis of customer data \cite{fin_usecase2_crosssell_health_insurance}. \href{https://ieeexplore.ieee.org/abstract/document/10475046}{Link}

\item TabPFN was used in corporate bond recovery rate prediction for credit risk management \cite{fin_usecase3_recovery_rate_github}. \href{https://github.com/hoanguyen94/Recovery-rate-prediction}{Link}

\end{enumerate}

\section*{Energy and Utilities}

We collected 15 use cases focused on environmental forecasting (algal blooms, wildfire, rainfall), renewable‑energy nowcasting, process/asset optimization across water, oil \& gas, and materials.
    
\begin{enumerate}

\item TabPFN was employed to predict river algal blooms through multi-classification of chlorophyll-a concentrations, aiding water management \cite{energy_usecase1_river_algal_tabpfn}. \href{https://koreascience.kr/article/JAKO202427157640711.page}{Link}

\item TabPFN facilitated wildfire propagation prediction in Canadian conifer forests, classifying fire types for environmental risk assessment \cite{energy_usecase2_wildfire_automl}. \href{https://www.sciencedirect.com/science/article/pii/S157495412400253X}{Link}

\item TabPFN was integrated into a machine learning framework for optimizing energy consumption at wastewater treatment plants \cite{energy_usecase3_wwtp_tabpfnreg}. \href{https://www.researchgate.net/publication/390516459_Machine_learning_framework_for_energy_consumption_optimization_using_the_TabPFNRegressor_algorithm}{Link}

\item TabPFN supported rainfall forecast post-processing using historical error patterns from environmental data \cite{energy_usecase4_rainfall_tabpfn}. \href{https://github.com/aarxshi/rainfall_tabpfn}{Link}

\item TabPFN enabled solar forecast error adjustment, particularly during rapid weather changes, as developed by Open Climate Fix \cite{energy_usecase5_solar_adjuster_ocf}. \href{https://gist.github.com/anshulg954/5f4423ee6b3d3151fa8d0d7fcd98d3eb}{Link}

\item TabPFN was applied to predict ash fusibility in high-alkali coal for improved energy production \cite{energy_usecase6_ash_fusibility_high_alkali}. \href{https://papers.ssrn.com/sol3/papers.cfm?abstract_id=5406504}{Link}

\item TabPFN contributed to predicting Henry coefficients for alkanes in zeolites, aiding hydroisomerization in sustainable fuel production \cite{energy_usecase7_henry_zeolites}. \href{https://pubs.acs.org/doi/full/10.1021/acs.jpcc.5c03868}{Link}

\item TabPFN facilitated shape-selectivity modeling in zeolites for long-chain alkane hydroisomerization, optimizing catalyst design \cite{energy_usecase8_shape_selectivity_zeolites}. \href{https://doi.org/10.4233/uuid:f36da034-5cb3-42ca-a53d-d351f68a9ffa}{Link}

\item TabPFN was used in an integrated framework for estimated ultimate recovery prediction and fracturing optimization in shale gas reservoirs \cite{energy_usecase9_shale_eur_fracturing}. \href{https://www.researchgate.net/publication/395761327_Coupling_EUR_Prediction_with_Fracturing_Optimization_An_Integrated_Machine_Learning_Framework_for_Shale_Gas_Development}{Link}

\item TabPFN supported core data augmentation for enhanced reservoir parameter prediction in oil and gas exploration \cite{energy_usecase10_core_augmentation_reservoir}. \href{https://www.researchgate.net/publication/395434405_Enhancing_Reservoir_Parameter_Prediction_Workflows_via_Advanced_Core_Data_Augmentation}{Link}

\item TabPFN was employed to optimize energy performance in multistage centrifugal pumps through entropy generation analysis \cite{energy_usecase11_multistage_pump_tabpfn}. \href{https://www.sciencedirect.com/science/article/abs/pii/S0360544225040411}{Link}

\item TabPFN contributed to physics-informed regression for evaluating solar-reflective materials in facade temperature modeling \cite{energy_usecase12_physinf_facade}. \href{https://arxiv.org/pdf/2507.16174}{Link}

\item TabPFN was applied to generate advanced global heat flow maps at 0.2° resolution, integrating high-resolution geophysical data to improve geothermal resource modeling \cite{energy_usecase13_global_heatflow_02}. \href{https://www.researchgate.net/publication/396728153_The_First_02_Resolution_Global_Continental_Heat_Flow_Map_Advancing_Fine-Scale_Geothermal_Modeling}{Link}

\item TabPFN contributed to FuelCast, standardizing benchmarks for ship fuel consumption prediction and improving efficiency in maritime operations \cite{energy_usecase14_fuelcast}. \href{https://arxiv.org/abs/2510.08217}{Link}

\item TabPFN was used as the main supervised classifier to automatically identify thunderstorm ground enhancements from particle detector and environmental measurements \cite{energy_usecase15_tge_tabpfn}. \href{https://arxiv.org/abs/2510.25125}{Link}

\end{enumerate}

\section*{Manufacturing and Industrial}

We collected 12 diverse use cases including anomaly detection, predictive maintenance, physics‑aware optimization—spanning IIoT security, rotating machinery, semiconductor testing, geotechnical/optical sensing, machining, battery thermal modeling, and concrete mix design.

\begin{enumerate}

\item TabPFN enabled early fault classification in rotating machinery, addressing data scarcity in industrial scenarios \cite{manuf_usecase1_rotating_faults_tabpfn}. \href{https://ieeexplore.ieee.org/abstract/document/10318062}{Link}

\item TabPFN facilitated microcontroller performance prediction, aiding semiconductor screening with minimal supervision, as studied at Infineon Technologies \cite{manuf_usecase2_mcu_performance_tabpfn}. \href{https://iris.polito.it/handle/11583/3002056}{Link}

\item TabPFN was applied to caisson inclination prediction in ultra-deep construction, combining data denoising techniques \cite{manuf_usecase3_caisson_inclination_ml}. \href{https://www.sciencedirect.com/science/article/abs/pii/S2214391225001734}{Link}

\item TabPFN supported event classification in phase-sensitive optical time-domain reflectometry systems for distributed fiber sensing \cite{manuf_usecase4_photdr_event_classification}. \href{https://opg.optica.org/oe/fulltext.cfm?uri=oe-33-17-36646&id=575783}{Link}

\item TabPFN was integrated into an adaptive ensemble for intrusion detection in Industrial Internet of Things networks \cite{manuf_usecase5_wfetab_iiot_ids}. \href{https://rdcu.be/eASzJ}{Link}

\item TabPFN enabled a random forest-based framework for attack recognition in Internet of Things networks, improving interpretability \cite{manuf_usecase6_rf_tabpfn_iot_attack}. \href{https://ieeexplore.ieee.org/stamp/stamp.jsp?tp=&arnumber=11142329}{Link}

\item TabPFN facilitated geotechnical site characterization for predicting soil strength and imputing mechanical parameters \cite{manuf_usecase7_geotech_site_char_tabpfn}. \href{https://arxiv.org/abs/2509.03191}{Link}

\item TabPFN was used in cryogenic-assisted abrasive waterjet machining for improving surface integrity in titanium alloys \cite{manuf_usecase8_cryo_awj_ti64}. \href{https://www.sciencedirect.com/science/article/abs/pii/S2214993725004531}{Link}

\item TabPFN supported in-context learning for thermal behavior prediction in nano-phase change materials for battery systems \cite{manuf_usecase9_nano_pcm_thermal_icl}. \href{https://www.sciencedirect.com/science/article/pii/S036054422504335X}{Link}

\item TabPFN was applied to explainable strength evaluation in multicomponent concrete mixtures \cite{manuf_usecase10_multicomponent_concrete}. \href{https://www.mdpi.com/1996-1944/18/19/4456}{Link}

\item TabPFN was integrated into a multimodal fusion framework linking microstructure to friction behavior in martensitic stainless steel, improving wear resistance in materials engineering applications \cite{manuf_usecase11_martensitic_friction_multimodal}. \href{https://papers.ssrn.com/sol3/papers.cfm?abstract_id=5346149}{Link}

\item TabPFN supported multiscale modeling to predict soil salinity in arid farmland, advancing sustainable agricultural management in regions such as Xinjiang \cite{manuf_usecase12_soil_salinity_multiscale}. \href{https://papers.ssrn.com/sol3/papers.cfm?abstract_id=5591702}{Link}

\end{enumerate}

\section*{Other Industries}

We collected 19 further  heterogeneous TabPFN applications spanning geoscience, forensic science, agriculture, materials, and engineering domains—ranging from microbiome classification and lunar regolith analysis to soil property modeling, crop yield and phenology forecasting, fuel-blend optimization, and spatial regression.
    
\begin{enumerate}

\item TabPFN was modified for microbiome data classification in metagenomics, matching species abundance patterns with synthetic priors \cite{other_usecase1_microbiome_zero_inflated}. \href{https://openreview.net/forum?id=3I0bVvUj25}{Link}

\item TabPFN enabled lunar regolith analysis for classifying meteorite compositions from spectral data \cite{other_usecase2_lunar_meteorites}. \href{https://www.sciencedirect.com/science/article/pii/S2095268624001010}{Link}

\item TabPFN facilitated winter wheat yield forecasting in agricultural regions by integrating climate and remote sensing data \cite{other_usecase3_winter_wheat_yield_ssrn}. \href{https://papers.ssrn.com/sol3/papers.cfm?abstract_id=5380177}{Link}

\item TabPFN was applied to flood impact assessment on housing prices by geographic areas \cite{other_usecase4_flood_housing_prices_ml_climate}. \href{https://github.com/melina-thegarza/ml-climate/blob/main/doc/ML_Climate___Final.pdf}{Link}

\item TabPFN showed the strongest performance on 31 predictive soil modeling datasets containing 30 to 460 samples \cite{other_usecase5_soil_mapping_new_default}. \href{https://arxiv.org/abs/2508.09888}{Link}

\item TabPFN was applied to shallow natural gas hazard prediction in tunnel construction \cite{other_usecase6_shallow_gas_tunnel_tabpfn}. \href{https://www.sciencedirect.com/science/article/pii/S2590123025029366}{Link}

\item TabPFN supported automated feature engineering for energy consumption forecasting in domain-specific applications \cite{other_usecase7_autoenergy_feature_eng}. \href{https://www.sciencedirect.com/science/article/pii/S0950705125013413}{Link}

\item TabPFN enabled Australian rice phenology prediction using remote sensing and weather data for crop management \cite{other_usecase8_rice_phenology_tabpfn}. \href{https://www.mdpi.com/2072-4292/17/17/3050}{Link}

\item TabPFN was applied to a multi-stage framework for predicting fuel blend properties through automated feature engineering \cite{other_usecase9_fuel_blend_framework}. \href{https://chemrxiv.org/engage/chemrxiv/article-details/68dc888d3e708a7649ff0ec9}{Link}

\item TabPFN enabled kriging prior regression for incorporating spatial context in soil mapping predictions \cite{other_usecase10_kriging_prior_regression}. \href{https://arxiv.org/abs/2509.09408}{Link}

\item TabPFN was applied to predicting electric vehicle crash severity using deep learning models \cite{other_usecase11_ev_crash_severity}. \href{https://www.arxiv.org/abs/2509.11449}{Link}

\item TabPFN enhanced clone-type recognition across programming languages through metrics-driven analysis, improving stability and interpretability in software engineering \cite{other_usecase12_clone_type}. \href{https://wiley.authorea.com/users/980519/articles/1346750-metrics-first-language-aware-clone-type-recognition-auditable-signals-across-c-c-java-and-python}{Link}

\item TabPFN was used to predict biomass-derived hard carbon performance in sodium-ion batteries, facilitating material selection for energy storage systems \cite{other_usecase13_hard_carbon_sib}. \href{https://arxiv.org/abs/2510.12833}{Link}

\item TabPFN informed the development of TabImpute, enabling efficient zero-shot imputation for missing tabular data and improving preprocessing pipelines \cite{other_usecase14_tabimpute}. \href{https://www.arxiv.org/abs/2510.02625}{Link}

\item TabPFN, alongside TabICL and related foundation models, was evaluated for intrusion detection, improving cybersecurity performance in IoT networks \cite{other_usecase16_cyber_fm_tabpfn_tabicl}. \href{https://www.mdpi.com/2079-9292/14/19/3792}{Link}

\item TabPFN supported continual learning for tabular data streams in resource-constrained environments \cite{other_usecase17_imlp_continual}. \href{https://arxiv.org/html/2510.04660v1}{Link}

\item TabPFN contributed to assessing robustness of language models for data fitting under irrelevant variations \cite{other_usecase19_llm_data_fitting_robustness}. \href{https://arxiv.org/pdf/2508.19563}{Link}

\item TabPFN was used in forensic science to advance biogeographical ancestry predictions \cite{Heinzel2025}. \href{https://www.sciencedirect.com/science/article/pii/S1872497325000705}{Link}

\item TabPFN was used as a benchmark model for predicting avocado alternate bearing from Sentinel-2 and climate features \cite{other_usecase21_avocado_alt_bearing}. \href{https://www.preprints.org/manuscript/202510.2413}{Link}

\end{enumerate}

\section{Data Contamination and Deduplication for Real-TabPFN-2.5}
\label{data_contamination_real_tabpfn}
To ensure fair evaluation and eliminate data contamination, we implemented an enhanced multi-tiered deduplication and filtering pipeline for Real-TabPFN-2.5. While based on the methodology used for Real-TabPFN \cite{garg2025realtabpfn}, the process was extended to deduplicate the training datasets against all internal benchmarks, our curated in-house validation suite, and the public TabArena benchmark \cite{erickson2025tabarena}. Our deduplication procedure combines automated cross-referencing of dataset identifiers, feature schemas, and row- and column-level hashes with manual metadata inspection to ensure that no training dataset overlaps with, or is derived from, any evaluation dataset. Datasets failing these criteria were excluded from the final training corpus.

\subsection{Training Datasets}
\label{appx.Training-Datasets}
The following table lists the datasets curated for fine-tuning, along with their sources and access links.

\begin{longtable}{p{10cm}p{4cm}}
\toprule
\textbf{Name} & \textbf{Source} \\
\midrule
\endfirsthead
\toprule
\textbf{Name} & \textbf{Source} \\
\midrule
\endhead

\href{https://www.openml.org/search?type=data&sort=runs&status=active&id=1459}{artificial-characters} & OpenML \\
\href{https://www.openml.org/search?type=data&status=active&id=251}{BNG(breast-w)} & OpenML \\
\href{https://www.openml.org/search?type=data&status=active&id=137}{BNG(tic-tac-toe)} & OpenML \\
\href{https://www.openml.org/d/40668}{connect\_4} & OpenML \\
\href{https://www.openml.org/search?type=data&sort=runs&status=active&id=1471}{eeg-eye-state} & OpenML \\
\href{https://openml.org/search?type=data&status=active&id=43551}{Employee-Turnover-at-TECHCO} & OpenML \\
\href{https://openml.org/search?type=data&status=active&id=1044}{eye\_movements} & OpenML \\
\href{https://www.openml.org/search?type=data&status=active&id=41787&sort=runs}{FOREX\_eurpln-hour-High} & OpenML \\
\href{https://www.openml.org/search?type=data&sort=runs&status=active&id=1476}{gas-drift} & OpenML \\
\href{https://openml.org/search?type=data&status=active&id=23512}{higgs} & OpenML \\
\href{https://openml.org/search?type=data&status=active&id=44201}{Intersectional-Bias-Assessment-(Training-Data)} & OpenML \\
\href{https://openml.org/search?type=data&status=active&id=43904}{law-school-admission-binary} & OpenML \\
\href{https://openml.org/search?type=data&status=active&id=43617}{Medical-Appointment} & OpenML \\
\href{https://www.openml.org/search?type=data&status=active&id=41671&sort=runs}{microaggregation2} & OpenML \\
\href{https://www.openml.org/search?type=data&sort=runs&id=901&status=active}{fried} & OpenML \\
\href{https://www.openml.org/search?type=data&status=active&id=43923&sort=runs}{mushroom} & OpenML \\
\href{https://openml.org/search?type=data&status=active&id=44226}{NewspaperChurn} & OpenML \\
\href{https://openml.org/search?type=data&status=active&id=1568}{nursery} & OpenML \\
\href{https://www.openml.org/search?type=data&status=active&id=46676&sort=runs}{WBCAtt} & OpenML \\
\href{https://www.openml.org/search?type=data&sort=runs&id=43039&status=active}{Internet Firewall Data} & OpenML \\
\href{https://www.kaggle.com/datasets/himselfthedecker/aam-avaliacao-dataset}{aam\_avaliacao\_dataset} & Kaggle \\
\href{https://www.kaggle.com/datasets/rohanshetty678/air-traffic-data}{Air Traffic Data} & Kaggle \\
\href{https://www.kaggle.com/datasets/stefadp/ansibledefectsprediction}{ansible-defects-prediction} & Kaggle \\
\href{https://www.kaggle.com/datasets/nehaprabhavalkar/av-healthcare-analytics-ii}{AV Healthcare Analytics II} & Kaggle \\
\href{https://www.kaggle.com/datasets/tarunchilkur/client}{Candidate Selection} & Kaggle \\
\href{https://www.kaggle.com/datasets/sulianova/cardiovascular-disease-dataset}{Cardio Disease} & Kaggle \\
\href{https://www.kaggle.com/datasets/aniketng21600/crop-damage-information-in-india}{Classification - Crop Damages in India (2015-2019)} & Kaggle \\
\href{https://www.kaggle.com/datasets/christianlillelund/csgo-round-winner-classification}{CSGO Round Winner Classification} & Kaggle \\
\href{https://www.kaggle.com/datasets/vpkprasanna/flower-type-prediction-machine-hack}{Flower Type Prediction Machine Hack} & Kaggle \\
\href{https://www.kaggle.com/datasets/gunner38/horseracing/data}{Horse Racing - Tipster Bets} & Kaggle \\
\href{https://www.kaggle.com/datasets/kanuriviveknag/road-accidents-severity-dataset}{How severe the accident could be} & Kaggle \\
\href{https://www.kaggle.com/datasets/pankeshpatel/hrcommasep}{hr-comma-sep} & Kaggle \\
\href{https://www.kaggle.com/datasets/jsrojas/ip-network-traffic-flows-labeled-with-87-apps}{ip-network-traffic-flows-labeled-with-87-apps} & Kaggle \\
\href{https://www.kaggle.com/datasets/pawan2905/jantahack-cross-sell-prediction}{Janatahack cross-sell prediction} & Kaggle \\
\href{https://www.kaggle.com/datasets/mamtadhaker/lt-vehicle-loan-default-prediction}{L\&T Vehicle Loan Default Prediction} & Kaggle \\
\href{https://www.kaggle.com/datasets/benfattori/league-of-legends-diamond-games-first-15-minutes}{League of Legends Diamond Games (First 15 Minutes)} & Kaggle \\
\href{https://www.kaggle.com/code/franciscoescobar/richter-s-predictor-modeling-earthquake-damage}{Richter's Predictor Modeling Earthquake Damage} & Kaggle \\
\href{https://www.kaggle.com/datasets/kartikjaspal/server-logs-suspicious}{Server Logs - Suspicious} & Kaggle \\
\href{https://www.kaggle.com/datasets/lucidlenn/sloan-digital-sky-survey}{Sloan Digital Sky Survey DR14} & Kaggle \\
\href{https://www.kaggle.com/datasets/muhakabartay/sloan-digital-sky-survey-dr16}{Sloan Digital Sky Survey DR16} & Kaggle \\
\href{https://www.kaggle.com/datasets/brajeshmohapatra/term-deposit-prediction-data-set}{Term Deposit Prediction Data Set} & Kaggle \\
\href{https://www.kaggle.com/datasets/danielamigo/trajectorybasedshipclassification/data}{trajectory-based-ship-classification} & Kaggle \\
\href{https://www.kaggle.com/datasets/mhdzahier/travel-insurance}{Travel Insurance} & Kaggle \\

\bottomrule
\end{longtable}

\section{Details on Causal Inference Results}
\label{sec:causal_inference}

\paragraph{Causal Inference} Most real-world decision problems ultimately hinge on causal questions—understanding what would happen if we intervened, rather than merely observing correlations. Estimating Conditional Average Treatment Effects (CATEs) is one of the central ways to answer these “what-if” questions: how would an individual’s outcome change if a treatment were applied versus withheld?

\paragraph{Unconfounded Settings.} Many causal inference methods require \textit{unconfoundedness}, which broadly states that there are no features not included in the dataset that influence both the \textit{treatment} variable and the outcome \cite{rosenbaum_propensity}. While recent studies have begun to challenge the validity and verifiability of this assumption \cite{karlsson_unconf, robertson_dopfn}, there are presently a wide variety of causal inference methods designed for the unconfounded setting \cite{curth_doing_great, oprescu_econml}.

\paragraph{Importance of Base Model.} Recent empirical findings have shown that when unconfoundedness holds, CATE estimation can be framed as an AutoML problem \cite{vanderschueren_autocate}, as many CATE estimators require a choice of classification or regression model to approximate the likelihood (propensity) of a treatment and an outcome given an individual's features. Parallel studies \cite{zhang_one_model, robertson_dopfn} have shown that TabPFN is an especially strong choice for meta-learners such as the X-, T-, and S-Learner \cite{kunzel_meta_learners}, hypothesizing that its strong performance in tabular prediction transfers to the problem of causal inference.

\begin{table}[h]
    \centering
    \caption{Description of causal inference datasets in the RealCause benchmark.}
    \label{tab:realcause_data}
    \begin{tabular}{l c c c c}
        \toprule
        \textbf{Characteristic} & \textbf{ACIC-2016} & \textbf{IHDP} & \textbf{Lalonde-CPS} & \textbf{Lalonde-PSID} \\
        \midrule
        Realizations & 10 & 100 & 100 & 100 \\
        Samples & 4,802 & 747 & 16,177 & 2,675 \\
        Features & 58 & 25 & 8 & 8 \\
        \bottomrule
    \end{tabular}
\end{table}

\section{The TabPFN Ecosystem}

\begin{figure}[h]
  \centering
  \includegraphics[width=\linewidth]{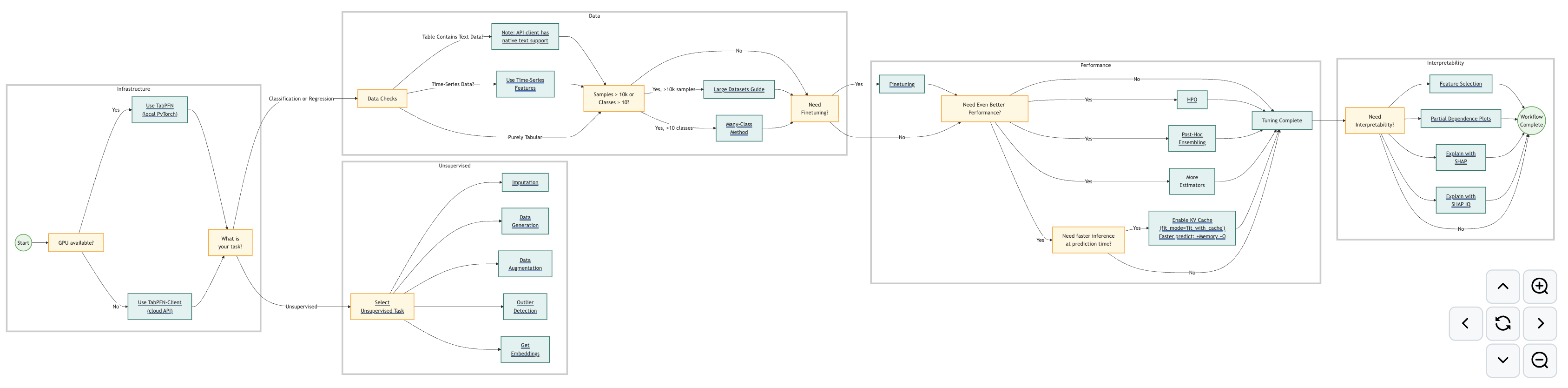}
  \caption{A minimal user workflow through components in the TabPFN–Extensions ecosystem.}
  \label{fig:workflow}
\end{figure}

Figure \ref{fig:workflow} provides a minimal user workflow through components in the TabPFN–Extensions ecosystem.

\section{Additional Internal Benchmark Details}
\label{app:additional-internal-results}

\subsection{Details on the normalization}
For benchmarking, we normalize scores per dataset to enable averaging and clearer comparison across datasets, ensuring that differences in dataset difficulty do not bias comparisons. For each dataset, we linearly scale scores between 0 (worse model on this dataset) and 1 (best model). For each model, the default and tuned versions are considered as two different models for the normalization. Bar heights show the mean normalized performance, and error bars denote the standard error of the mean (SEM) across datasets, reflecting uncertainty from dataset variability. 

\subsection{Additional results on many features}

\begin{figure}[htbp]
\centering
\begin{minipage}[t]{.48\textwidth}
    \centering
    \includegraphics[scale=0.65]{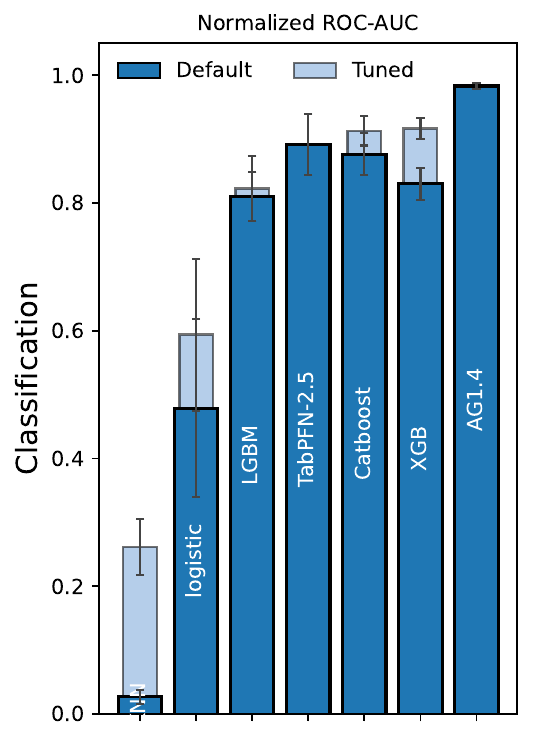}
    \end{minipage} 
\begin{minipage}[t]{.48\textwidth}
    \centering
    \includegraphics[scale=0.65]{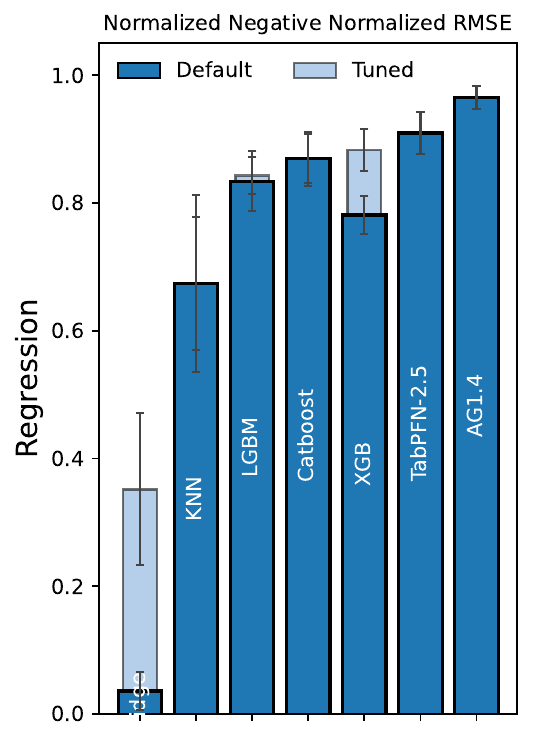}
\end{minipage}
\caption{\textbf{TabPFN-2.5 default performs well up to 2{,}000 features}. In our internal benchmark on datasets from 500 features to 2{,}000 features, we can see that for both classification (left) and regression (right), the default TabPFN-2.5 outperforms any other default model and is better than any tuned single model for regression.}
\label{fig:internal-large-feature-split-results}
\end{figure}

In Figure \ref{fig:internal-large-feature-split-results}, we show results on an internal set of datasets containing from 500 to 2,000 features showing strong default performance.

\section{Detailed TabArena Results}
\label{app:tabarena-detailed}
\begin{figure}[htbp]
\centering
\begin{minipage}[t]{.48\textwidth}
  \centering
  \includegraphics[trim={0cm 0 2.6cm 0},clip,scale=0.38]{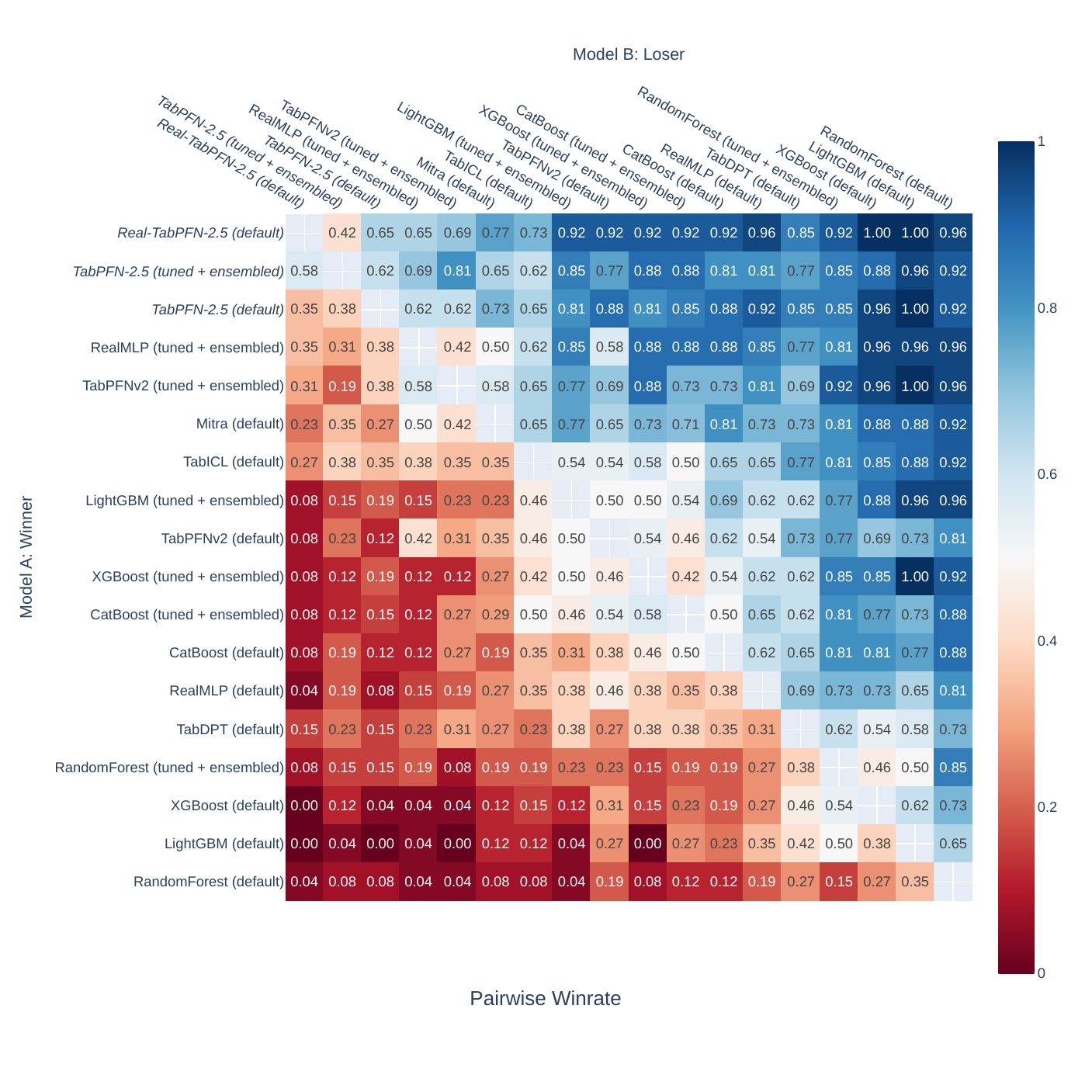}
\end{minipage}%
\hfill
\begin{minipage}[t]{.48\textwidth}
  \centering
  \includegraphics[trim={1.6cm 0 0cm 0},clip,scale=0.38]{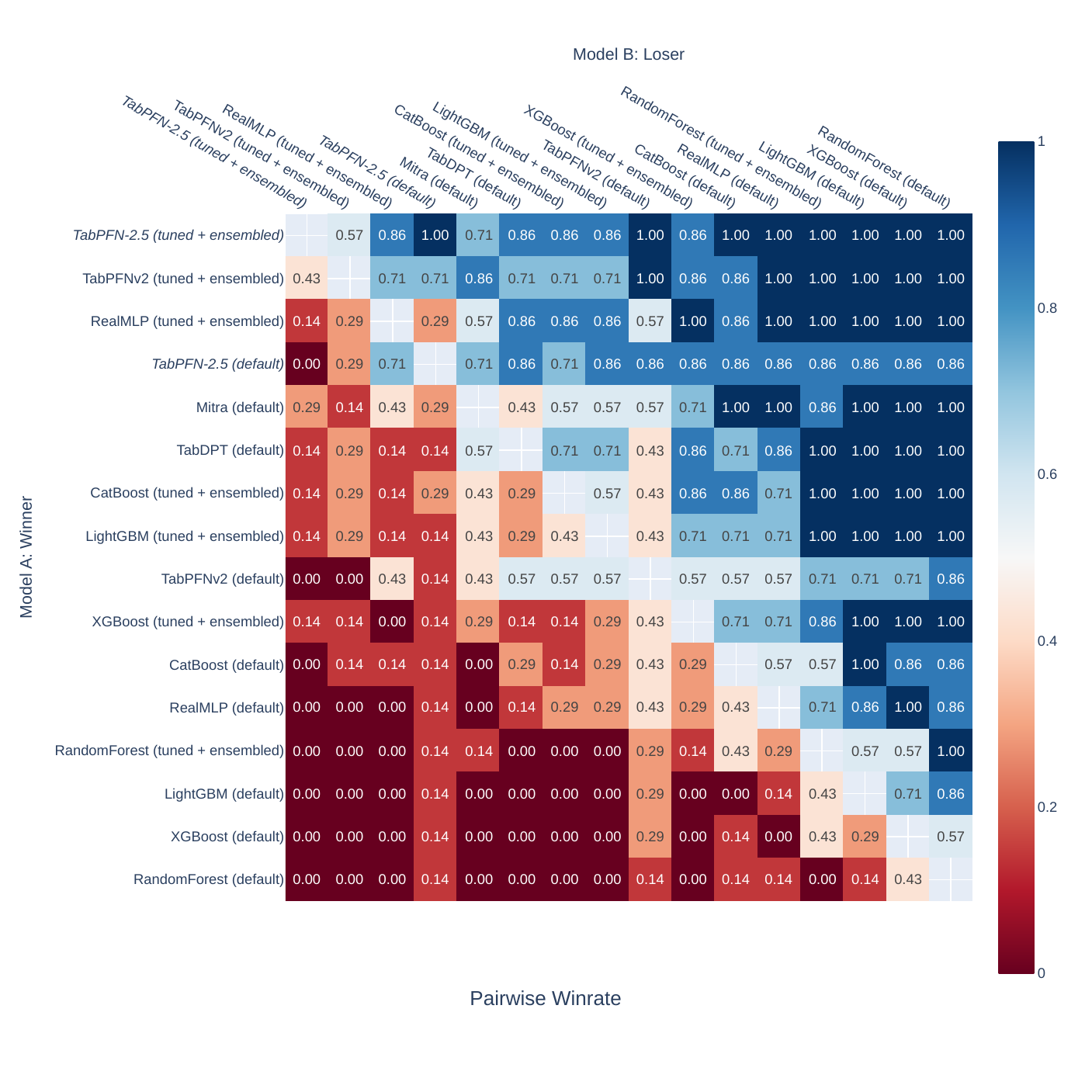}
\end{minipage}
\caption{TabArena-Lite pairwise win rates on \textbf{classification} (left) and \textbf{regression} (right), restricted to TabPFNv2 compatible datasets (less than \textbf{10K training samples and 500 features}). Note that tuning for TabPFN-2.5 is only based on 60 random configs compared to 200 for the baselines.}
\label{fig:tabarena-winrates-10k-500}
\end{figure}

\begin{figure}[htbp]
\centering
\begin{minipage}[t]{.48\textwidth}
  \centering
  \includegraphics[trim={0cm 0 2.6cm 0},clip,scale=0.38]{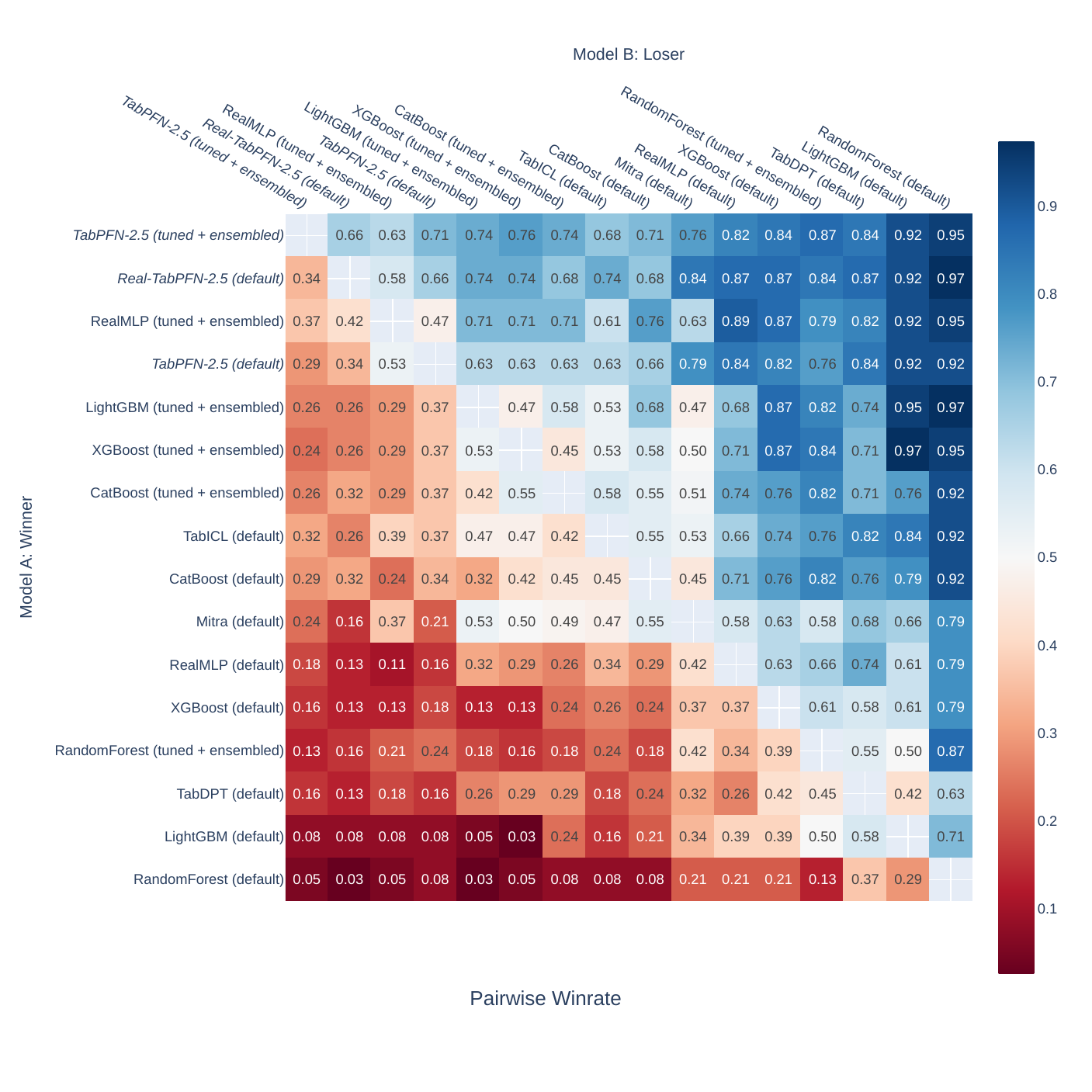}
\end{minipage}%
\hfill
\begin{minipage}[t]{.48\textwidth}
  \centering
  \includegraphics[trim={1.6cm 0 0cm 0},clip,scale=0.38]{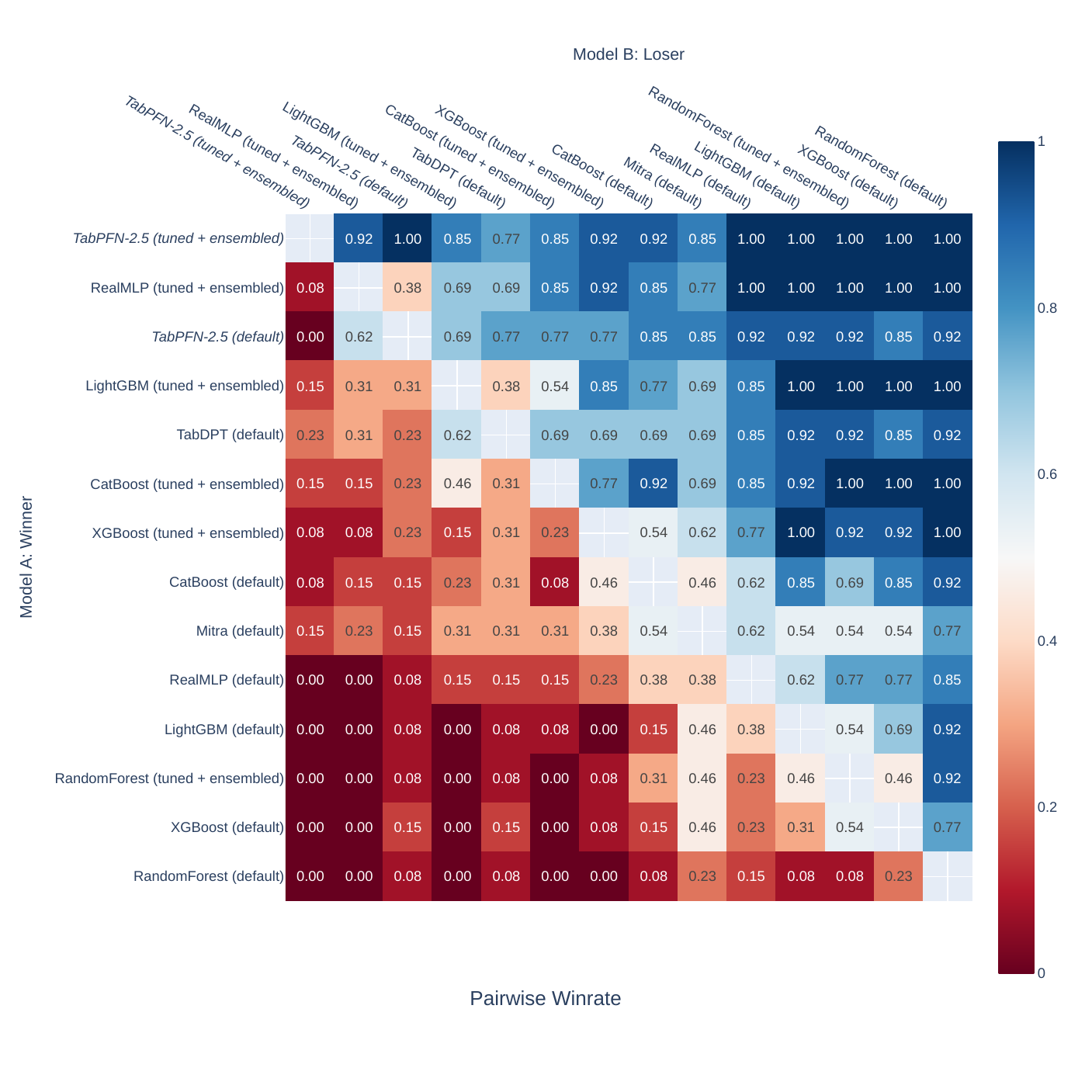}
\end{minipage}
\caption{TabArena-Lite pairwise win rates on \textbf{classification} (left) and \textbf{regression} (right), evaluated on \textbf{all datasets} (up to \textbf{100k training samples and 2K features}). Note that tuning for TabPFN-2.5 is only based on 60 random configs compared to 200 for the baselines.}
\label{fig:tabarena-winrates-100k-2k}
\end{figure}

In addition to the results shown in Section \ref{sec:results}, we also report the pairwise winrates of different models on TabArena in Figure \ref{fig:tabarena-winrates-10k-500} (for TabPFNv2 compatible datasets with less than 10k rows and 500 features) and Figure \ref{fig:tabarena-winrates-100k-2k} (all datasets up to 100k training rows and 2k features).

\begin{figure}[htbp]
    \centering
    \includegraphics[width=\linewidth]{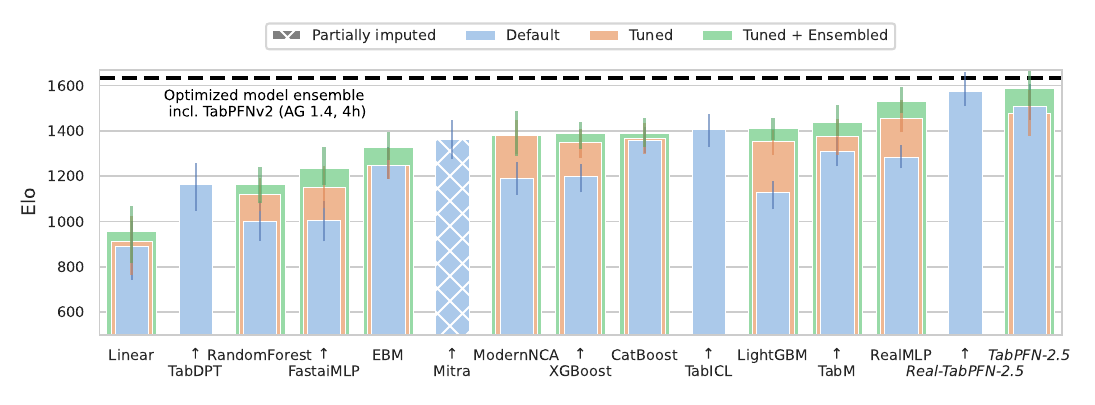}
    \caption{\textbf{Comparison with TabICL \citep{qu2025tabicl}}. In this plot, we show the performance of TabPFN-2.5 and TabICL on a TabArena-lite subset compatible with TabICL, restricting to  \textbf{classification datasets with less than 500 features}. On this subset for which TabICL is designed, we see that TabPFN-2.5 significantly outperforms TabICL.}
    \label{fig:tabarena-tabicl}
\end{figure}

\begin{figure}[htbp]
    \centering
    \includegraphics[width=\linewidth]{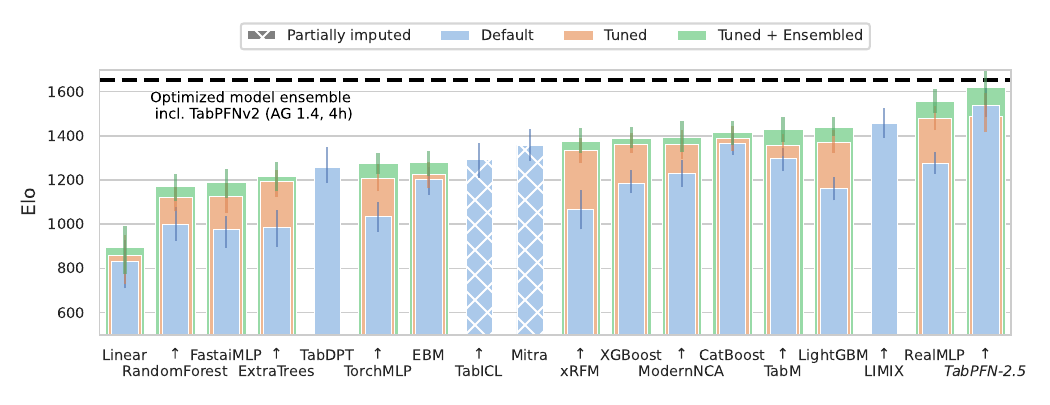}
    \caption{\textbf{Comparison with LimiX \citep{zhang2025limix}}. In this plot, we show the performance of TabPFN-2.5 and LimiX on datasets from TabArena-Lite with \textbf{less than 50{,}000 training samples and less than 2{,}000 features}, which corresponds to the datasets on which the TabArena maintainers could run LimiX at the time of writing (see this \href{https://github.com/autogluon/tabarena/pull/208}{link}). On this subset, we see that TabPFN-2.5 significantly outperforms LimiX. Note that these results are still unverified by the original authors at the time of writing and thus not included in the main paper results.}
    \label{fig:tabarena-limix}
\end{figure}

We also compare our TabPFN-2.5 model to other foundation models in more detail below. In Figure \ref{fig:tabarena-tabicl}, we show that TabPFN-2.5 outperforms TabICL when we restrict TabArena to only datasets for which TabICL is designed, and in Figure \ref{fig:tabarena-limix}, we show much better performance when compared to LimiX's results on datasets with less than 50{,}000 samples and 2{,}000 features, which corresponds to the datasets on which the TabArena maintainers could run LimiX at the time of writing (see this \href{https://github.com/autogluon/tabarena/pull/208}{link}).

\section{Results with Tuned Decision Thresholds}
\label{app:threshold_tuning}

\begin{figure}[H]
\centering
   \includegraphics[scale=0.5]{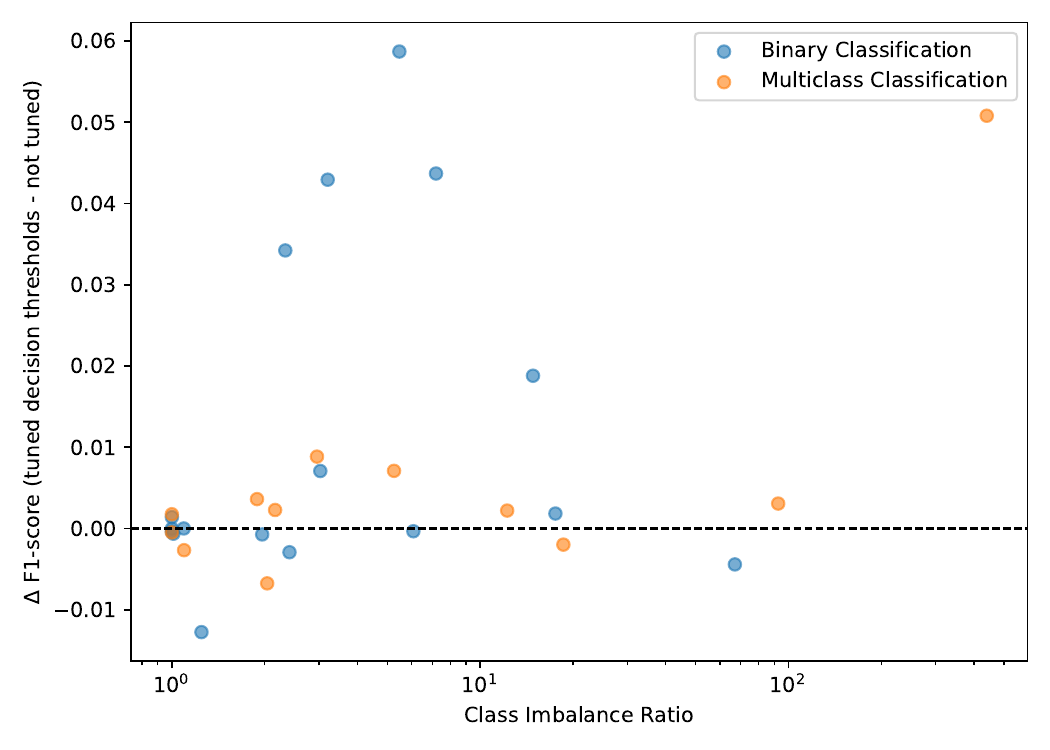}
  \caption{\textbf{F1-score sometimes improves substantially by decision threshold tuning.} The plot shows the difference in F1-score (macro) between a model with an optimized decision threshold and the same model using a default (untuned) threshold. This demonstrates the effectiveness of the tuning procedure for metric-specific optimization.}
  \label{fig:tuned_F1_delta}
\end{figure}

Starting with TabPFN-2.5, our framework supports tuning the decision threshold to optimize for specific metrics. Figure \ref{fig:tuned_F1_delta} quantifies the performance gains that this procedure can yield, illustrating substantial improvement in F1-score for several imbalanced datasets when tuning the threshold.

\FloatBarrier

\section{Supplementary Inference Time Details}

\begin{figure}
    \centering
    \includegraphics{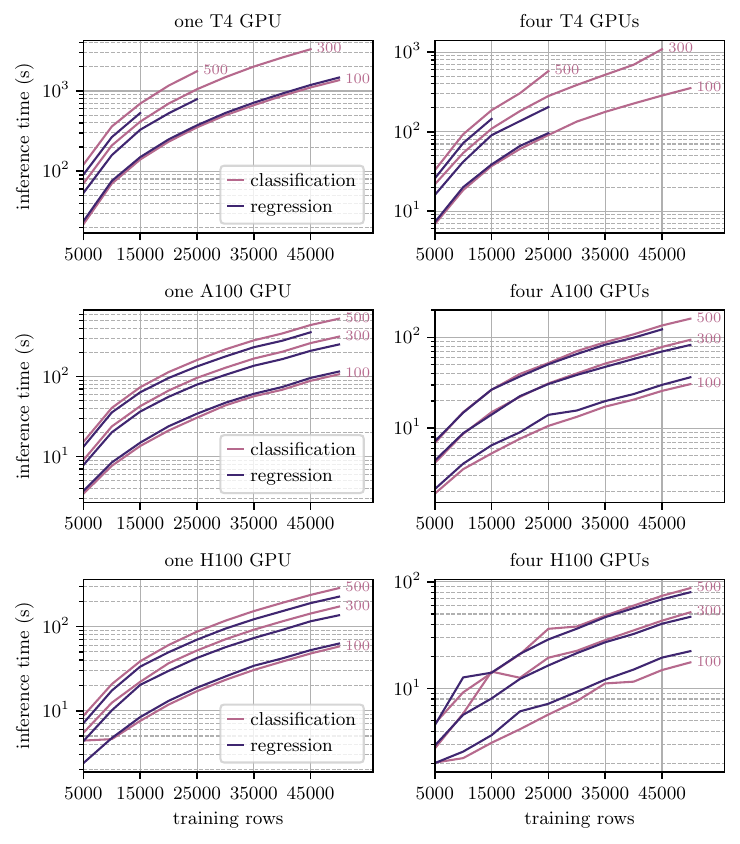}
    \caption{
        Time taken, in seconds, to train TabPFN-2.5 models on various training set sizes, and then make predictions on $500$ test rows, using three common models of NVIDIA GPU: T4 15GB, A100 SXM 40GB, H100 SXM 80GB.
        Performance is shown for $100$, $300$, and $500$ features.
        Datasets with more than $500$ features have the same performance as datasets with $500$, as each estimator will subsample to $500$ features.
        Incomplete lines indicate that the GPU had insufficient memory for that dataset size.
    }
    \label{fig:inference-time-by-gpu}
\end{figure}

\begin{figure}[htbp]
    \centering
    \includegraphics{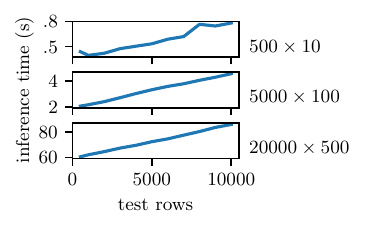}
    \caption{
        The time taken by TabPFN-2.5 to train and predict scales linearly in the test set size, shown here for a classification model trained on datasets of $500$ rows $\times$ 10 features, 5{,}000 rows $\times$ 100 features, and 20{,}000 rows $\times$ 500 features.
        Measured on one H100 GPU.
    }
    \label{fig:inference-time-test-size}
\end{figure}

\begin{figure}[htbp]
    \centering
    \includegraphics[scale=0.53]{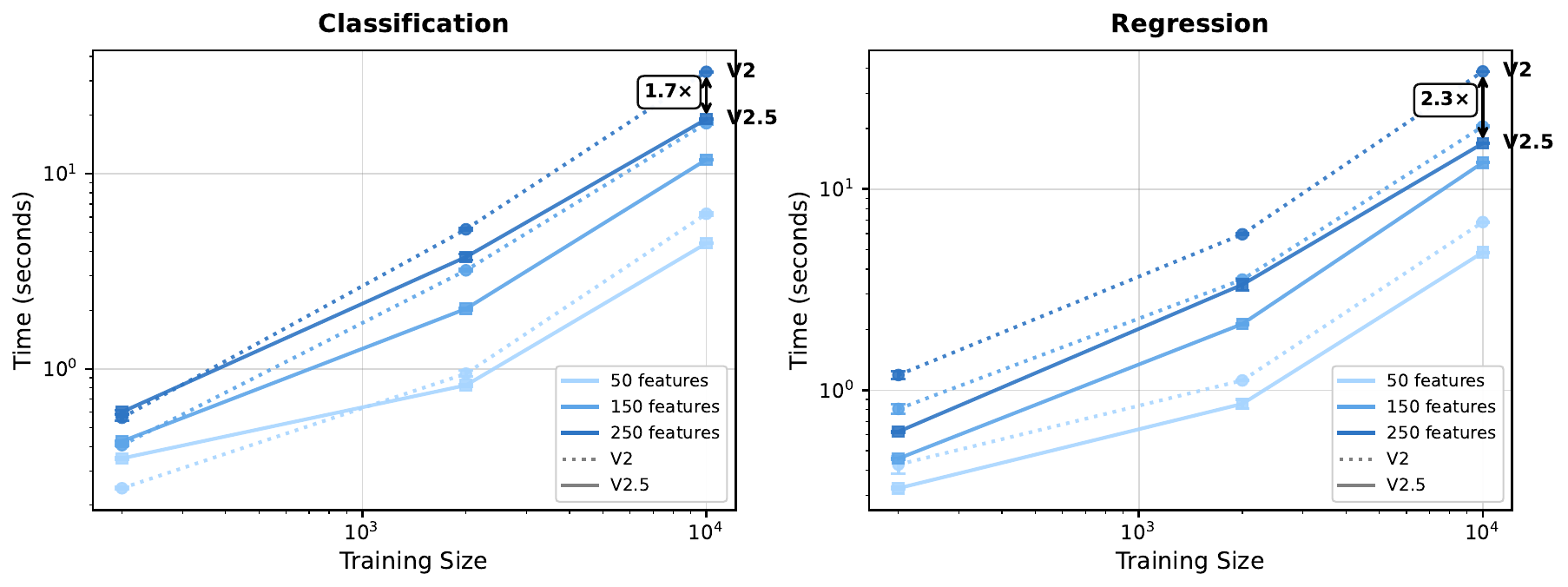}
    \caption{
        \textbf{TabPFN-2.5 is significantly faster than TabPFNv2}. Comparison of the time taken to fit + predict TabPFN-2.5 vs TabPFNv2 on different number of rows and features. Measured for 100 test points on 1 H100, using the same number of estimators (8). Note that this is measured using the v2 and v2.5 versions available on the latest release of the TabPFN package, and thus is on top of the performance improvements since the original release of TabPFNv2.
    }
    \label{fig:v2-vs-v25-time}
\end{figure}

Figure~\ref{fig:inference-time-by-gpu} shows the inference latency you can expect for three common models of GPUs.
Figure~\ref{fig:inference-time-test-size} shows that the time scales linearly with the number of test rows.
Figure~\ref{fig:v2-vs-v25-time} compares the fit + training time of TabPFN-2.5 vs TabPFNv2, showing that TabPFN-2.5 is significantly faster, showing between 1x and 2.3x speedup depending on the dataset size.


\end{document}